\documentclass{article}

\PassOptionsToPackage{numbers, compress}{natbib}
\usepackage[main, final]{neurips_2026}

\usepackage{microtype}
\usepackage{graphicx}
\usepackage{subcaption}
\usepackage{booktabs} % for professional tables
\usepackage{hyperref}
\usepackage{amsmath}
\usepackage{amssymb}
\usepackage{mathtools}
\usepackage{amsthm}
\usepackage{graphicx}
\usepackage{threeparttable}
\usepackage{rotating}
\usepackage{multirow}
\usepackage{diagbox}
\usepackage{colortbl}
\usepackage{subcaption}
\usepackage{wrapfig}
\usepackage{bm}
\usepackage{mathtools}
\usepackage{pifont}
\newcommand{\xmark}{\ding{55}} % 엑스 표시

\usepackage[capitalize,noabbrev]{cleveref}

\theoremstyle{plain}

\theoremstyle{definition}

\theoremstyle{remark}

\usepackage[textsize=tiny]{todonotes}

\title{Discretizing Continuous Time Series for Imputation \\with Masked Diffusion Training}

\author{%
  Dongbin Kim\textsuperscript{1},\quad Seungyun Lee\textsuperscript{1},\quad Geonwoo Shin\textsuperscript{1},\quad Jaewook Lee\textsuperscript{1 *} \\[1ex]
  \textsuperscript{1}Seoul National University \\[1ex]
  \texttt{\{dongbin413,rats96,shin0621,jaewook\}@snu.ac.kr}
}

\begin{document}

\maketitle

\renewcommand{\thefootnote}{\fnsymbol{footnote}}
\footnotetext[1]{Corresponding author.}
\renewcommand{\thefootnote}{\arabic{footnote}}
\setcounter{footnote}{0}

\begin{abstract}
Time series imputation is a crucial area for reliable time series analysis, yet it remains challenging due to the complex temporal dynamics and noise of real-world data. Existing approaches, however, exhibit two limitations: missing and observed values are embedded within the same representation space without explicit structural separation, and continuous diffusion-based methods recover the missing values only through many iterative denoising steps rather than by predicting them directly. To address these, we propose the Masked Diffusion Time-series Imputation Model (MDTIM), which leverages the training paradigm of masked diffusion model for imputation tasks. The \texttt{[MASK]} token is structurally orthogonal to valid observations, and because the corrupted state is the partially observed series itself, the model is trained to predict the missing values directly from the observed ones, naturally aligning both the representation and the learning objective with the imputation task. To bridge the gap between discrete masked diffusion and the continuous, ordinal nature of time series, we further introduce Stochastic Discretization, which maps continuous values to ordinal-aware tokens while preserving continuous dynamics. Our experiments on diverse benchmarks confirm that MDTIM achieves superior robustness and scalability, consistently outperforming state-of-the-art deterministic and generative baselines across various missing scenarios. Code and project page are available at \url{https://dongbeank.github.io/MDTIM/}.
\end{abstract}

\section{Introduction}
\label{sec:intro}

Time series imputation is a fundamental task in time series analysis, as real-world data are often partially observed due to sensor failures or transmission errors \citep{kim2023probabilistic}. At its core, imputation aims to recover the underlying clean signal from partial observations, which requires the representation and learning objective of the model to be well-aligned with the structure of partial observability.

From this perspective, we observe that existing approaches leave room for improvement along two axes. First, at the representation level, discriminative Transformer-based methods~\cite{du2023saits} and continuous diffusion models~\cite{tashiro2021csdi} inform the model of missing positions through a missingness indicator, an attention mask, or a conditioning mask, and the question is how that information enters the network. An indicator marks the position, but a placeholder value (e.g., zero) still occupies the same embedding space as valid observations, so the network has to learn to discount it~\cite{fadlon2025a}. An attention mask can remove an interaction, but it cannot announce anything to the rest of the sequence. Second, at the objective level, continuous diffusion models such as CSDI~\cite{tashiro2021csdi} are trained to denoise a Gaussian-corrupted copy of the missing values at a sampled noise level, so the quantity imputation actually needs, the posterior over the missing values given the observed ones, is obtained only after the full reverse chain is integrated.

% From this perspective, we observe that existing approaches leave room for improvement along two axes. First, at the representation level, both discriminative Transformer-based methods \citep{du2023saits} and continuous diffusion models \citep{tashiro2021csdi} embed missing and observed values within the same data manifold. Although attention masks can prevent the model from attending to missing positions, placeholder values (e.g., zeros) still occupy the same embedding space as valid observations, potentially introducing spurious correlations in the input representation \citep{fadlon2025a}. Second, at the objective level, continuous diffusion models such as CSDI \citep{tashiro2021csdi} are typically trained to predict the Gaussian noise added to the data rather than the original values themselves, which is somewhat indirect for a task whose goal is to recover what was originally there.

Masked Diffusion Models (MDM), recently developed in Natural Language Processing~\cite{sahoo2024simple}, offer a natural fit for both aspects. Unlike continuous diffusion that corrupts data with Gaussian noise, MDM replaces tokens with a special \texttt{[MASK]} symbol and learns to recover the original values. The \texttt{[MASK]} token is structurally orthogonal to any valid observation, and it is a symbol that the rest of the sequence reads, so every observed position sees where the gaps are and conditions on their shape. Because the corrupted state is the partially observed series itself, the model is trained to predict the clean value at every masked position given the observed ones, and a single forward pass returns the full distribution over every missing value, which is exactly the quantity imputation asks for. We compare these mechanisms under controlled conditions in Section~\ref{sec:ablation}.

% Masked Diffusion Models (MDM), recently developed in Natural Language Processing \citep{sahoo2024simple}, offer a natural fit for both aspects. Unlike continuous diffusion that corrupts data with Gaussian noise, MDM replaces tokens with a special \texttt{[MASK]} symbol and learns to recover the original values. The \texttt{[MASK]} token is structurally orthogonal to any valid observation, which provides a clearer separation between missing and observed states in the representation. The model also directly predicts the original values rather than the noise, more closely matching the imputation goal.

Despite this conceptual fit, directly applying MDM to time series imputation presents a fundamental challenge. Time series data are inherently continuous and ordinal, whereas language tokens are discrete and categorical. Naive quantization severs the ordinal relationships between adjacent values and loses information below the grid resolution, which is particularly problematic when fine-grained reconstruction is needed. This raises our central research question: \textit{How can we leverage the training paradigm of masked diffusion while preserving the continuous and ordinal nature of time series data?}

\begin{figure}[t]
    \centering
    \includegraphics[width=\linewidth]{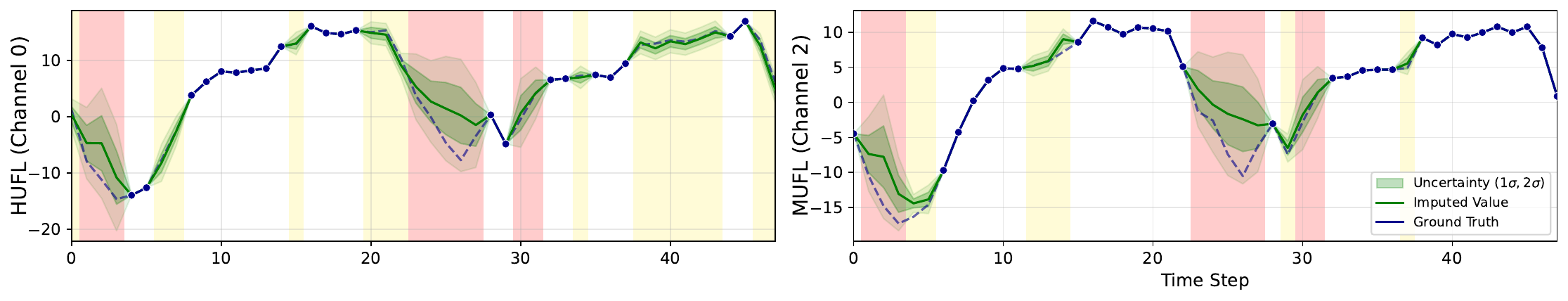}
    \caption{\textbf{Imputation results of MDTIM on the ETTh dataset.}
    Red regions indicate intervals where both channels are missing, while yellow regions represent single-channel missingness.
    The uncertainty ($\sigma$), derived from the output probability of MDTIM, is estimated to be higher in the red regions \textbf{(information scarcity)} compared to the yellow regions \textbf{(richer information)}.}
    \label{fig:imputation_intro}
\end{figure}

To answer this, we propose the \textbf{Masked Diffusion Time-series Imputation Model (MDTIM)}, a framework that bridges discrete masked diffusion and continuous time series modeling.
We introduce \textit{Stochastic Discretization}, which injects noise during tokenization to preserve information in expectation, and replace the standard one-hot objective with \textit{Ordinal-Aware Soft Labeling} to capture the ordinal relationships between tokens.
We further incorporate a spectral consistency objective and an expectation-based decoder for precise continuous reconstruction.

Our main contributions are summarized as follows:
\begin{itemize}
    \item We propose a Masked Diffusion Framework for time series imputation in which missingness enters the model as a symbol the sequence reads, giving a structural separation between missing and observed representations.
    \item We introduce a discretization pipeline adapted to masked diffusion imputation, combining \textit{Stochastic Discretization} and \textit{Ordinal-Aware Soft Labeling}, which preserves ordinal relationships and is unbiased in expectation.
    \item We employ an Expectation-based Unmasking strategy and Spectral Consistency Regularization, allowing MDTIM to reconstruct continuous time series precisely. Our method consistently outperforms state-of-the-art deterministic and generative baselines across diverse benchmarks.
\end{itemize}

\section{Related Work}

\subsection{Time Series Imputation}

Time series data are often partially observed due to sensor faults, irregular sampling, or data-collection constraints \citep{10.24963/ijcai.2025/1187}.
Naive deletion or mean/zero imputation can bias estimation and degrade downstream tasks \citep{kim2023probabilistic}, so time-series imputation aims to recover missing values by leveraging temporal dependencies, cross-variable correlations, and informative missingness patterns \citep{che2018recurrent, du2023saits}.

Methodologically, deep learning approaches have progressed from RNN-based architectures with masking \citep{che2018recurrent, cao2018brits} to Transformer-based models that leverage self-attention to capture long-range dependencies; SAITS \citep{du2023saits}, in particular, achieves strong performance via Diagonally-Masked Self-Attention.
Beyond deterministic models, diffusion-based models such as CSDI \cite{tashiro2021csdi} and SSSD \cite{lopez2023diffusion} produce probabilistic imputations by conditioning on observed data and iterative denoising. SSSD further incorporates structured state space models to better capture long-range temporal dependencies.
%However, this conditional formulation is misaligned with the imputation task: the model learns \textit{``what noise was added"} rather than \textit{``what was originally there"}.
%Moreover, its training objective is decoupled from the masking ratio, applying uniform weight regardless of reconstruction difficulty.
In this formulation an absent value cannot be represented by Gaussian corruption, so missingness has to be supplied from outside as a mask, and the missing values are recovered only after the full reverse chain. In masked diffusion, erasure is itself missingness, so a partially observed series is already a valid state of the forward process and the model fits $p_\theta(x_0 \mid x_t)$ directly.

\subsection{Masked Diffusion Model}

Discrete diffusion models extend the diffusion paradigm to categorical state spaces \citep{hoogeboom2021argmax}.
Among them, D3PM \citep{austin2021structured} introduced an absorbing-state formulation, where tokens are progressively replaced by a special \texttt{[MASK]} symbol and recovered through iterative denoising.
MDLM \citep{sahoo2024simple} adapted this framework to language modeling via a continuous-time weighted cross-entropy objective, demonstrating its effectiveness as a non-autoregressive alternative for conditional text generation.
Building on MDLM, subsequent works have improved robustness through unmasking-order-aware training \citep{kim2025train} and conditional fidelity by treating observed tokens as fixed anchors during sampling \citep{koh2025conditional, rout2025anchored}.
Despite these advances, masked diffusion has been studied almost exclusively in discrete modalities such as NLP, and its extension to time series remains underexplored due to the fundamental gap between discrete tokens and continuous, ordinal temporal signals.

\begin{figure}[t!]
    \centering
    \includegraphics[width=\textwidth]{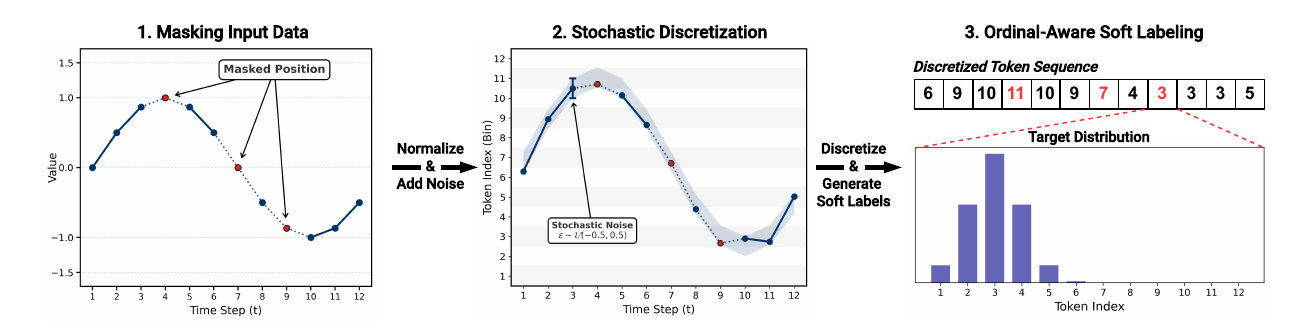} 
    
    \caption{\textbf{Overview of the proposed Stochastic Discretization and Ordinal-Aware Soft Labeling framework.} 
    The process consists of three stages: 
    (1) \textit{Masking Input Data}, where missing values are identified; 
    (2) \textit{Stochastic Discretization}, which applies instance-adaptive normalization and injects stochastic noise $\epsilon \sim \mathcal{U}(-0.5, 0.5)$ to bridge the continuous-discrete gap; and 
    (3) \textit{Ordinal-Aware Soft Labeling}, which maps continuous values to discrete tokens while generating soft target distributions to preserve ordinal semantic relationships.}
    
    \label{fig:stochastic_discretization}
\end{figure}

\section{Discrete Representation for Time-Series Modeling}
\label{sec:discrete_rep}

Although the Masked Diffusion Model (MDM) offers a promising framework for imputation, directly applying it to time series presents a structural challenge: MDMs are inherently designed for discrete state spaces (e.g., text tokens), whereas time series data consist of continuous numerical values. 
To bridge this gap, we introduce a unified discretization framework illustrated in Figure~\ref{fig:stochastic_discretization}, which combines \textit{Stochastic Discretization} to mitigate quantization error and \textit{Ordinal-Aware Soft Labeling} to preserve the temporal order of the original signals within the discrete vocabulary. 

\subsection{Stochastic Discretization}
\label{sec:stochastic_tokenization}

\paragraph{Instance-Adaptive Normalization.} To stabilize local statistics against distribution shifts, we normalize each input window $x \in \mathbb{R}^{T \times C}$ into $\tilde{x}$. Specifically, the normalization statistics are computed solely using the observed values (excluding masked positions), ensuring that there is no data-leakage in the training and imputation process.

\paragraph{Stochastic Token Generation.} We map the normalized values to a discrete vocabulary of size $K$. Specifically, we first project the continuous value $\tilde{x}_{t,c}$ onto the grid coordinates and inject uniform noise $\epsilon \sim \mathcal{U}(-0.5, 0.5)$ before rounding:
\begin{equation}
    c_{t,c} = \frac{\tilde{x}_{t,c} - v_{\min}}{v_{\max} - v_{\min}} \cdot (K - 1) + 1, \quad 
    z_{t,c} = \lfloor c_{t,c} + \epsilon \rceil,
    \label{eq:stochastic_tokenization}
\end{equation}
where $\lfloor \cdot \rceil$ denotes the nearest integer function. This stochastic mechanism preserves information in expectation. For instance, a coordinate $c_{t,c} = 3.6$ is assigned to token $4$ with $60\%$ probability and token $3$ with $40\%$, ensuring $\mathbb{E}[z_{t,c}] \approx c_{t,c}$. This allows the model to learn precise dynamics beyond the fixed grid resolution.

\begin{figure}[t!]
    \centering
    \includegraphics[width=1\textwidth]{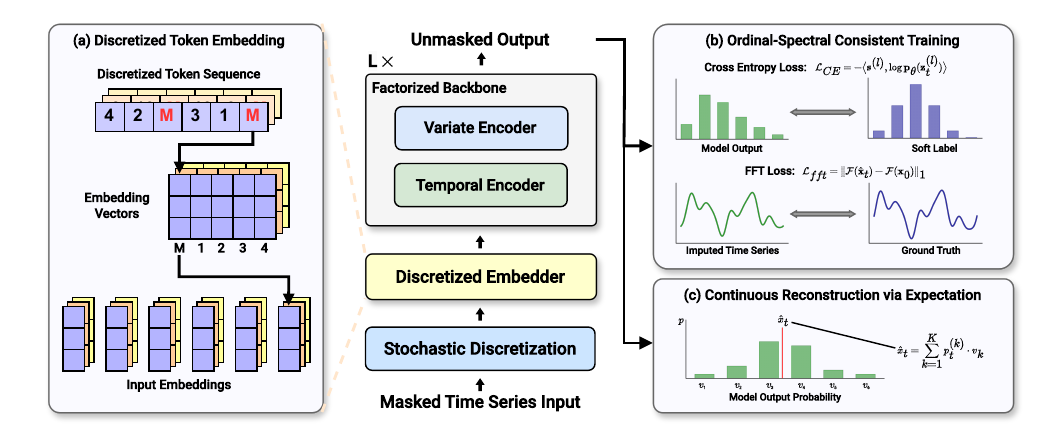}
    \caption{
        \textbf{Overview of the Masked Diffusion Time-series Imputation Model (MDTIM) framework.}
        The architecture operates on the discretized input through three key phases:
        \textbf{(a) Discretized Token Embedding:} The discrete tokens generated by the stochastic tokenizer are mapped into dense vector representations.
        \textbf{(b) Ordinal-Spectral Consistent Training:} The backbone is trained with a dual objective combining Ordinal-Aware Soft Cross-Entropy and Spectral Consistency Loss. This ensures spectral consistency across global frequencies while maintaining local ordinal accuracy.
        \textbf{(c) Continuous Reconstruction via Expectation:} Finally, continuous values are recovered by computing the probability-weighted expectation of the predicted token distribution, enabling precise dense reconstruction from discrete outputs.
    }
    \label{fig:MDTIM_framework}
\end{figure}

\subsection{Ordinal-Aware Soft Labeling}
\label{sec:soft_labeling}

Treating discretized tokens as independent classes ignores their ordinal nature (e.g., token $k$ is semantically closer to $k+1$ than $k+10$). To enforce continuity, we employ Ordinal-Aware Soft Labeling with a truncated Gaussian kernel.

Given a ground-truth token index $y \in \{1, \dots, K\}$, the target probability $s_i$ for the $i$-th class is defined as:
\begin{equation}
    s_i = \begin{cases} 
    \frac{1}{Z} \exp\left(-\frac{(i-y)^2}{\sigma^2}\right) & \text{if }  |i-y| \le w, \\
    0 & \text{otherwise},
    \end{cases}
    \label{eq:soft_label}
\end{equation}
where $w$ is the truncation window size, $\sigma$ controls the smoothness in units of token bins, and $Z$ is the normalization constant. We use $w = 2$ and $\sigma = 1$ for every dataset and missing rate, fixed a priori (Appendix~\ref{app:hyper}). Crucially, we assign zero probability to the \texttt{[MASK]} token ($s_0 = 0$) and distant bins. This concentrates probability mass on valid ordinal neighbors while preserving the structural separation of the mask state.

\section{Masked Diffusion Framework}
\label{sec:framework}

Building upon the discrete representation of time series data described in Section~\ref{sec:discrete_rep}, we introduce the \textbf{Masked Diffusion Time-series Imputation Model (MDTIM)}, a framework designed to model the joint distribution of multivariate time series and incorporate training techniques of masked diffusion models. Figure~\ref{fig:MDTIM_framework} illustrates the overall pipeline.

\subsection{Factorized Temporal-Variate Backbone}
To effectively model the joint distribution of multivariate time series, we employ a Factorized Temporal-Variate Transformer based on the Diffusion Transformer (DiT) proposed by \citet{peebles2023scalable}. This architecture processes the input through alternating factorization of the time ($T$) and feature ($C$) axes.

\paragraph{Discretized Embedding.}
Given the masked input indices $\mathbf{z}_t \in \{0, \dots, K\}^{B \times T \times C}$ at diffusion step $t$, we utilize a distinct embedding matrix $\mathbf{E}_c$ for each channel $c$ to preserve semantic orthogonality. The input is projected to an initial hidden state $\mathbf{H}^0 \in \mathbb{R}^{B \times T \times C \times D}$, where $D$ is the hidden dimension.
The \texttt{[MASK]} token (index $0$) receives its own embedding, so masked positions are present as keys in every attention block and observed positions can read where the gaps are, rather than merely being prevented from attending to them.

\paragraph{Interleaved DiT Blocks.}
The core network consists of $L$ layers, each sequentially processing the temporal and feature axes via two DiT blocks conditioned on the diffusion timestep $t$, with Rotary Position Embeddings (RoPE)~\cite{su2024roformer} encoding relative positions.
The Temporal block reshapes $\mathbf{H}^{(l-1)}$ to $\mathbb{R}^{(B \cdot C) \times T \times D}$ and captures time-dependencies along the temporal axis:
\begin{equation}
    \mathbf{H}'^{(l)} = \text{TemporalEncoder}(\mathbf{H}^{(l-1)}, t).
\end{equation}
The Variate block then reshapes $\mathbf{H}'^{(l)}$ to $\mathbb{R}^{(B \cdot T) \times C \times D}$ and symmetrically models cross-channel correlations:
\begin{equation}
    \mathbf{H}^{(l)} = \text{VariateEncoder}(\mathbf{H}'^{(l)}, t).
\end{equation}
The final output is projected to vocabulary size $K_{\text{out}} = 1.5K$, covering the extended range $[-1.5, 1.5]$. Since normalization uses the observed entries only, a missing value can fall outside the range they span, and the wider output grid lets the model place it there. The soft label of Eq.~\ref{eq:soft_label} keeps its mass within $w$ bins of the true token, so such extreme predictions stay extreme rather than drifting toward the center of the grid.

\subsection{Ordinal-Spectral Consistent Training Objective}
To simultaneously ensure local reconstruction fidelity and global temporal coherence, we formulate a dual-domain objective. This combines a discrete diffusion loss adapted for ordinal continuity with an auxiliary spectral regularization term.

\paragraph{Ordinal-Aware Masked Diffusion.}
We adopt the continuous-time training paradigm of Masked Diffusion Language Models (MDLM)~\cite{austin2021structured, sahoo2024simple}, optimizing a weighted variational lower bound (NELBO). Standard MDLM employs a simple cross-entropy loss against one-hot targets, which treats all incorrect tokens equally, failing to capture the ordinal magnitude inherent in time-series data.

To address this, we employ the \textit{Ordinal-Aware Soft Labels} derived in Sec.~\ref{sec:soft_labeling} as the optimization targets. For a set of masked indices $\mathcal{M}_t$ at timestep $t$, the diffusion loss is formulated as:
\begin{equation}
    \mathcal{L}_{\text{diff}} = \mathbb{E}_{t} \left[ w(t) \sum_{l \in \mathcal{M}_t} - \big\langle \mathbf{s}^{(l)}, \log \mathbf{p}_\theta(\mathbf{z}_t^{(l)}) \big\rangle \right],
    \label{eq:diff_loss}
\end{equation}
where $\log \mathbf{p}_\theta(\mathbf{z}_t^{(l)}) \in \mathbb{R}^K$ is the predicted log-probability vector, and $\mathbf{s}^{(l)} \in \mathbb{R}^K$ denotes the soft target distribution.
The detailed formulation of time-dependent weight $w(t)$ and the corresponding noise schedule are detailed in Appendix~\ref{noise}.

Replacing the one-hot target $\mathbf{e}_{y}$ with $\mathbf{s}^{(l)}$ can be viewed as a smooth relaxation of the standard MDLM objective.
Writing $\mathbf{s}^{(l)} = (1-\epsilon) \mathbf{e}_{y} + \epsilon\,\mathbf{u}^{(l)}$ with $\epsilon = 1 - s_y^{(l)}$, the per-token loss decomposes as:
\begin{equation}
    -\langle \mathbf{s}^{(l)}, \log \mathbf{p}_\theta \rangle 
    = (1-\epsilon) \mathcal{L}_{\mathrm{NELBO}}^{(l)}
    + \epsilon D_{\mathrm{KL}}(\mathbf{u}^{(l)} \Vert \mathbf{p}_\theta)
    + \mathrm{const},
    \label{eq:decomposition}
\end{equation}
so that the original NELBO is recovered when $\mathbf{s}^{(l)}$ collapses to a one-hot ($\epsilon \to 0$) and otherwise encourages the predicted distribution to align with the ordinal neighborhood of $y$.
The detailed derivation is provided in Appendix~\ref{app:soft_label}.

\paragraph{Spectral Consistency Regularization.}
While the token-wise objective ensures local accuracy, it may neglect global temporal correlations and frequency structures. To strictly enforce global coherence, we incorporate a \textit{Spectral Consistency Loss}, adapting the frequency-domain regularization proposed in DiffusionTS~\cite{yuan2024diffusionts}.
During training, we compute a continuous estimate of the full sequence, $\hat{\mathbf{x}}_t$, by applying expectation-based decoding (Sec.~\ref{sec:inference}). We then minimize the $L_1$ distance between the Fourier representations of the reconstructed and ground-truth signals:
\begin{equation}
    \mathcal{L}_{\text{FFT}} = \mathbb{E}_{t} \left[ \| \mathcal{F}(\hat{\mathbf{x}}_t) - \mathcal{F}(\mathbf{x}_0) \|_1 \right],
\end{equation}
where $\mathcal{F}(\cdot)$ denotes the Fast Fourier Transform (FFT).

Therefore, the total training objective is a weighted combination: $\mathcal{L}_{\text{total}} = \mathcal{L}_{\text{diff}} + \lambda \mathcal{L}_{\text{FFT}}$.

\subsection{Continuous Reconstruction via Expectation}
\label{sec:inference}

Since \textit{Stochastic Discretization} injects random noise into the input, the model output depends on the specific noise realization. To obtain a robust estimate, we run inference $M$ times (e.g., $M=10$) with independent noise samples and average the predicted probability distributions, $\bar{p}_{t,c} = \frac{1}{M} \sum_{m=1}^{M} p_{t,c}^{(m)}$. 
The continuous value is then recovered as the expectation over the averaged distribution:
\begin{equation}
    \hat{x}_{t,c} = \sum_{k=1}^{K} \bar{p}_{t,c}^{(k)} \cdot v_k,
    \label{eq:expectation}
\end{equation}
where $v_k$ is the center value of the $k$-th bin within $[-1.5, 1.5]$. The result is then denormalized using the instance-wise statistics to recover the original scale.

\section{Experiments}
\label{sec:experiments}

We empirically present a comprehensive evaluation of the proposed Masked Diffusion Time-series Imputation Model (MDTIM), comparing it against state-of-the-art deterministic and generative baselines. Our experiments focus on imputation accuracy, robustness to complex missing patterns, and computational efficiency.

\subsection{Experimental Setup}

We evaluate our method on widely used benchmarks: Energy, ETTh, Weather, and the synthetic Sine dataset. To rigorously test the models, we simulate data corruption using two distinct mechanisms: \textit{Uniform} masking, where observations are dropped randomly to mimic independent failures, and \textit{Geometric} masking, which removes consecutive time steps to simulate prolonged sensor malfunctions or transmission errors. For both scenarios, we assess performance across varying missing rates of 30\%, 50\%, and 70\%.

We compare MDTIM against a comprehensive suite of 19 baselines spanning various architectural paradigms. These include RNN-based methods such as BRITS\citep{cao2018brits}, MRNN\citep{yoon2018estimating}, and GRU-D\citep{che2018recurrent}; and Transformer-based models including the canonical Transformer\citep{vaswani2017attention}, Informer\citep{zhou2021informer}, PatchTST\citep{nie2023a}, SAITS\citep{du2023saits}, and ImputeFormer\citep{nie2024imputeformer}. Also, we include recent high-performance convolutional and linear architectures: TimesNet\citep{wu2023timesnet}, SCINet\citep{liu2022scinet}, DLinear\citep{zeng2023transformers}, FILM\citep{zhou2022film}, and FreTS\citep{yi2023frequency}. Finally, we evaluate against probabilistic frameworks including GP-VAE\citep{fortuin2020gp}, US-GAN\citep{miao2021generative}, and the diffusion-based CSDI\citep{tashiro2021csdi} and FGTI\citep{yang2024frequency}. We evaluate each trained model under three random mask seeds and report the averaged Mean Absolute Error (MAE). Variation across training seeds is reported in Appendix~\ref{app:seeds}.

\subsection{Imputation Performance}
\label{subsec:main_performance}

\paragraph{Main Results.}
As shown in Table~\ref{tab:main_results1}, MDTIM achieves the lowest MAE on Energy, ETTh, and Weather across both Uniform and Geometric missing patterns. FGTI~\citep{yang2024frequency}, a frequency-aware diffusion model, attains the best results on the synthetic Sine dataset whose signal is dominated by a few well-defined frequencies, while MDTIM trails by only a marginal gap (e.g., 0.002 vs.\ 0.001 at 30\% Uniform). On the more complex real-world ETTh dataset, however, FGTI degrades substantially (0.218 vs.\ 0.127 at 30\% Uniform), suggesting that frequency-domain priors generalize less reliably to irregular and non-stationary signals. Compared to the diffusion-based CSDI~\citep{tashiro2021csdi}, MDTIM consistently yields lower MAE across all datasets, owing to our \textit{Expectation-based Reconstruction} that recovers continuous values deterministically from the predicted token distribution and mitigates the sampling variance inherent in continuous diffusion. MDTIM also exhibits notable robustness under geometric masking, where most baselines suffer significant degradation; for instance, on Energy at 30\%, MDTIM increases only from 0.044 to 0.053, indicating that the masked diffusion objective encourages inference of global temporal structure rather than reliance on local interpolation.

\begin{table*}[t!]
\setlength{\tabcolsep}{3pt}
\centering
\renewcommand{\arraystretch}{1.13}
\caption{Quantitative comparison of multivariate time-series imputation performance ($L=48$). We report the MAE of MDTIM and baselines averaged over 3 random mask seeds at missing ratios of 30\% and 70\% (50\% in Appendix \ref{app:full_results}). The best results are highlighted in \textbf{bold}.}
\label{tab:main_results1}
\resizebox{1\columnwidth}{!}{%
\begin{tabular}{l|l|l|l|l|l|l|l|l|l|l|l|l|l|l|l|l|l}
\hline
\hline
\rowcolor{gray!20}
\multicolumn{2}{c|}{\textbf{Dataset}}
&\multicolumn{4}{c|}{\textbf{Energy}}&\multicolumn{4}{c|}{\textbf{ETTh}}&\multicolumn{4}{c|}{\textbf{Weather}}&\multicolumn{4}{c}{\textbf{Sine}}\\ 
\hline
\rowcolor{gray!20}
\multicolumn{2}{c|}{\textbf{Missing Type}}
&\multicolumn{2}{c|}{\textbf{Uniform}}&\multicolumn{2}{c|}{\textbf{Geometric}}&\multicolumn{2}{c|}{\textbf{Uniform}}&\multicolumn{2}{c|}{\textbf{Geometric}}&\multicolumn{2}{c|}{\textbf{Uniform}}&\multicolumn{2}{c|}{\textbf{Geometric}}&\multicolumn{2}{c|}{\textbf{Uniform}}&\multicolumn{2}{c}{\textbf{Geometric}}\\ 
\hline
\rowcolor{gray!20}
\multicolumn{2}{c|}{Model}
&30\%&70\%&30\%&70\%&30\%&70\%&30\%&70\%&30\%&70\%&30\%&70\%&30\%&70\%&30\%&70\%\\ 
\hline
\multirow{3}{*}{\begin{turn}{90}RNN\end{turn}}
 & BRITS                    & 0.246          & 0.379          & 0.329          & 0.366          & 0.194          & 0.299          & 0.227          & 0.292          & 0.050          & 0.073          & 0.058          & 0.071          & 0.010          & 0.021          & 0.020          & 0.019          \\
 & MRNN                     & 1.086          & 1.143          & 1.087          & 1.143          & 0.743          & 0.782          & 0.751          & 0.781          & 0.651          & 0.661          & 0.654          & 0.659          & 0.169          & 0.170          & 0.169          & 0.170          \\
 & GRUD                     & 0.364          & 0.457          & 0.426          & 0.445          & 0.310          & 0.394          & 0.348          & 0.383          & 0.104          & 0.369          & 0.164          & 0.350          & 0.008          & 0.014          & 0.013          & 0.012          \\ \hline
\multirow{5}{*}{\begin{turn}{90}Transformer\end{turn}}
 & Transformer              & 0.323          & 0.401          & 0.354          & 0.393          & 0.168          & 0.252          & 0.183          & 0.247          & 0.076          & 0.070          & 0.083          & 0.068          & 0.039          & 0.046          & 0.042          & 0.045          \\
 & Informer                 & 0.344          & 0.379          & 0.371          & 0.373          & 0.207          & 0.302          & 0.226          & 0.296          & 0.047          & 0.059          & 0.052          & 0.058          & 0.027          & 0.039          & 0.036          & 0.037          \\
 & PatchTST                 & 0.586          & 0.529          & 0.528          & 0.517          & 0.202          & 0.272          & 0.231          & 0.262          & 0.074          & 0.082          & 0.081          & 0.077          & 0.015          & 0.017          & 0.019          & 0.015          \\
 & SAITS                    & 0.177          & 0.212          & 0.204          & 0.208          & \underline{0.140} & \textbf{0.210}    & \underline{0.152}    & \underline{0.206} & 0.045          & 0.049          & 0.050          & 0.048          & 0.026          & 0.026          & 0.030          & 0.025          \\
 & ImputeFormer             & 0.066          & 0.219          & 0.113          & 0.194          & 0.146          & 0.245          & 0.165          & 0.235          & 0.049          & 0.171          & 0.090          & 0.149          & 0.006          & 0.041          & 0.012          & 0.037          \\
 \hline
\multirow{2}{*}{\begin{turn}{90}CNN\end{turn}}
 & TimesNet                 & 0.617          & 0.818          & 0.689          & 0.808          & 0.593          & 0.719          & 0.628          & 0.717          & 0.211          & 0.351          & 0.225          & 0.348          & 0.167          & 0.220          & 0.189          & 0.217          \\
 & SCINet                   & 0.557          & 0.571          & 0.546          & 0.565          & 0.238          & 0.323          & 0.261          & 0.317          & 0.074          & 0.089          & 0.086          & 0.087          & 0.019          & 0.031          & 0.026          & 0.029          \\
 \hline
\multirow{3}{*}{\begin{turn}{90}Linear\end{turn}}
 & DLinear                  & 0.795          & 0.527          & 0.684          & 0.517          & 0.379          & 0.445          & 0.373          & 0.435          & 0.372          & 0.223          & 0.345          & 0.220          & 0.070          & 0.053          & 0.062          & 0.051          \\
 & FiLM                     & 0.877          & 0.520          & 0.791          & 0.512          & 0.696          & 0.627          & 0.707          & 0.623          & 0.380          & 0.209          & 0.360          & 0.206          & 0.127          & 0.109          & 0.132          & 0.108          \\
 & FreTS                    & 0.157          & 0.219          & 0.225          & 0.206          & 0.222          & 0.299          & 0.262          & 0.286          & 0.071          & 0.080          & 0.080          & 0.074          & 0.068          & 0.068          & 0.075          & 0.065          \\  \hline
\multirow{5}{*}{\begin{turn}{90} Generative \end{turn}}
 & GPVAE                    & 0.476          & 0.730          & 0.501          & 0.727          & 0.333          & 0.449          & 0.369          & 0.440          & 0.158          & 0.255          & 0.172          & 0.252          & 0.152          & 0.162          & 0.153          & 0.162          \\
 & USGAN                    & 0.264          & 0.412          & 0.330          & 0.401          & 0.208          & 0.302          & 0.240          & 0.295          & 0.087          & 0.122          & 0.101          & 0.118          & 0.014          & 0.029          & 0.026          & 0.026          \\ 
 & CSDI                     & 0.094          & 0.139          & 0.110          & 0.136          & 0.160          & 0.244          & 0.177          & 0.240          & \underline{0.039}    & \underline{0.049}    & \underline{0.044}    & \underline{0.048}    & 0.003          & 0.004          & 0.004          & \underline{0.004} \\
 & FGTI                     & \underline{0.050}    & \underline{0.100}    & \underline{0.065}    & \underline{0.097}    & 0.218          & 0.349          & 0.293          & 0.328          & \underline{0.038}    & 0.051          & 0.046          & 0.049          & \textbf{0.001} & \textbf{0.003} & \textbf{0.002} & \textbf{0.003} \\
 & MDTIM (Ours)    & \textbf{0.044} & \textbf{0.085}          & \textbf{0.053} & \textbf{0.082}  & \textbf{0.127} & \underline{0.211} & \textbf{0.146}    & \textbf{0.205}  & \textbf{0.032} & \textbf{0.044}   & \textbf{0.036} & \textbf{0.043}  & \underline{0.002} & \underline{0.004} & \underline{0.003} & \textbf{0.003} 
 \\ \hline  \hline
\end{tabular} 
}
\end{table*}

\begin{table}[h!]
\setlength{\tabcolsep}{3pt}
\centering
\renewcommand{\arraystretch}{1.15}
\caption{Robustness analysis on extended sequence lengths ($L \in \{96, 192\}$). We report the MAE averaged over 3 random mask seeds.}
\label{tab:seq_len}
\resizebox{0.9\columnwidth}{!}{%
\begin{tabular}{l|c|c|c|c|c|c|c|c|c|c|c|c}
\hline
\hline
\rowcolor{gray!20}
\multicolumn{1}{c|}{\textbf{Length}}
&\multicolumn{6}{c|}{\textbf{96}}&\multicolumn{6}{c}{\textbf{192}}\\ 
\hline
\rowcolor{gray!20}
\multicolumn{1}{c|}{\textbf{Missing Type}}
&\multicolumn{3}{c|}{\textbf{Uniform}}&\multicolumn{3}{c|}{\textbf{Geometric}}&\multicolumn{3}{c|}{\textbf{Uniform}}&\multicolumn{3}{c}{\textbf{Geometric}}\\ 
\hline
\rowcolor{gray!20}
\multicolumn{1}{c|}{Model}
&30\%&50\%&70\%&30\%&50\%&70\%&30\%&50\%&70\%&30\%&50\%&70\%\\ 
\hline
BRITS                    & 0.191 & 0.229 & 0.287 & 0.220 & 0.250 & 0.280 & 0.192 & 0.230 & 0.287 & 0.222 & 0.250 & 0.280 \\
SAITS                    & \underline{0.139} & \underline{0.162} & \underline{0.204} & \underline{0.150} & \underline{0.172} & \underline{0.200} & \underline{0.145} & \underline{0.162} & \underline{0.198} & \underline{0.156} & \underline{0.171} & \underline{0.195} \\
CSDI                     & 0.149 & 0.177 & 0.222 & 0.162 & 0.187 & 0.218 & 0.170 & 0.199 & 0.245 & 0.183 & 0.209 & 0.241 \\
MDTIM                    & \textbf{0.124} & \textbf{0.152} & \textbf{0.199} & \textbf{0.138} & \textbf{0.163} & \textbf{0.194} & \textbf{0.123} & \textbf{0.148} & \textbf{0.190} & \textbf{0.135} & \textbf{0.157} & \textbf{0.186}
 \\ \hline  \hline
\end{tabular} 
}
\end{table}

\paragraph{Scalability to Long-Term Dependencies.}
To verify that MDTIM captures long-range temporal dependencies, we extend the evaluation to longer sequence lengths $L \in \{96, 192\}$. As shown in Table~\ref{tab:seq_len}, the performance advantage of MDTIM grows with the sequence length: while MDTIM is comparable to SAITS at $L=48$, the gap widens substantially at longer horizons. At $L=192$ with 30\% Uniform missing, MDTIM achieves an MAE of 0.123, considerably lower than BRITS (0.192) and SAITS (0.145), indicating that MDTIM captures global context without suffering from the attention dilution often observed in baselines.

\paragraph{Probabilistic Imputation Performance}

Beyond point estimation accuracy, we evaluate the quality of the predictive distributions generated by the models using the Continuous Ranked Probability Score (CRPS). CRPS estimates the calibration (reliability) and sharpness (precision) of the probabilistic inference.

\begin{table}[ht]
\centering
\caption{Quantitative comparison of probabilistic imputation performance (CRPS, $L=48$). We compare MDTIM with the diffusion-based baseline CSDI. Lower is better.}
\label{tab:crps_results}
\resizebox{0.8\columnwidth}{!}{%
\renewcommand{\arraystretch}{1.15}
\begin{tabular}{c|c|cc|cc|cc}
\hline
\hline
\rowcolor{gray!20}
\multicolumn{2}{c|}{} & \multicolumn{2}{c|}{\textbf{Energy}} & \multicolumn{2}{c|}{\textbf{ETTh}} & \multicolumn{2}{c}{\textbf{Sine}} \\
\hline
\rowcolor{gray!20}
\textbf{Missing Type} & \textbf{Rate} & CSDI & MDTIM & CSDI & MDTIM & CSDI & MDTIM \\
\hline
\multirow{3}{*}{Uniform}   & 30\% & 0.0629 & \textbf{0.0429} & 0.1120 & \textbf{0.0863} & 0.0023 & \textbf{0.0020} \\
                           & 50\% & 0.0747 & \textbf{0.0541} & 0.1356 & \textbf{0.1078} & 0.0025 & \textbf{0.0020} \\
                           & 70\% & 0.0927 & \textbf{0.0771} & 0.1748 & \textbf{0.1520} & 0.0030 & \textbf{0.0020} \\
\hline
\multirow{3}{*}{Geometric} & 30\% & 0.0732 & \textbf{0.0547} & 0.1262 & \textbf{0.1017} & 0.0027 & \textbf{0.0020} \\
                           & 50\% & 0.0818 & \textbf{0.0626} & 0.1460 & \textbf{0.1197} & 0.0028 & \textbf{0.0020} \\
                           & 70\% & 0.0905 & \textbf{0.0733} & 0.1708 & \textbf{0.1462} & 0.0029 & \textbf{0.0020} \\
\hline
\hline
\end{tabular}
}
\end{table}

As shown in Table~\ref{tab:crps_results}, MDTIM consistently outperforms CSDI, the diffusion-based baseline, across all datasets and missing scenarios.
On the highly periodic Sine dataset, MDTIM achieves a near-constant CRPS regardless of the missing rate, indicating that it captures deterministic periodicity with high confidence.
While continuous diffusion models often exhibit excessive variance due to sampling noise, MDTIM mitigates such ambiguity through its expectation-based decoding over discrete distributions.
Furthermore, on the complex real-world ETTh dataset, MDTIM achieves significantly lower CRPS (e.g., 0.0863 vs.\ 0.1120 at 30\% uniform missing), demonstrating that our \textit{Ordinal-Aware Soft Labeling} guides the model to produce well-calibrated uncertainty estimates.

\paragraph{Imputation under Naturally Missing Real-World Data.}

To assess MDTIM under realistic missingness patterns, we further evaluate on PhysioNet 2012~\citep{silva2012predicting}, a multivariate clinical time series benchmark with $\sim$80\% natural missingness due to irregular ICU sampling, and apply additional uniform and geometric masking on top of the existing missingness. Since about 80\% of entries are naturally missing, the observed values are too sparse to form a reliable spectrum of the target, so we disable the spectral consistency loss for this dataset; all other settings remain identical.

\begin{wraptable}{r}{0.5\columnwidth}
\vspace{-1.0em}
\centering
\caption{Imputation performance on PhysioNet 2012.}
\vspace{-0.5em}
\label{tab:physionet}
\setlength{\tabcolsep}{3pt}
\renewcommand{\arraystretch}{1.1}
\resizebox{0.5\columnwidth}{!}{%
\begin{tabular}{l|ccc|ccc}
\hline
\hline
\rowcolor{gray!20}
\textbf{PhysioNet2012} & \multicolumn{3}{c|}{\textbf{Uniform}} & \multicolumn{3}{c}{\textbf{Geometric}} \\
\hline
\rowcolor{gray!20}
\textbf{Model} & 30\% & 50\% & 70\% & 30\% & 50\% & 70\% \\
\hline
BRITS         & 0.333 & 0.364 & 0.408 & 0.356 & 0.379 & 0.405 \\
SAITS         & 0.283 & 0.314 & 0.360 & 0.298 & 0.324 & 0.358 \\
CSDI          & 0.303 & 0.332 & 0.376 & 0.320 & 0.342 & 0.374 \\
ImputeFormer  & 0.289 & 0.329 & 0.389 & 0.311 & 0.343 & 0.388 \\
FGTI          & 0.338 & 0.376 & 0.425 & 0.370 & 0.397 & 0.423 \\
MDTIM (Ours)  & \textbf{0.236} & \textbf{0.279} & \textbf{0.339} & \textbf{0.257} & \textbf{0.291} & \textbf{0.337} \\
\hline
\hline
\end{tabular}}
\vspace{-0.5em}
\end{wraptable}

As shown in Table~\ref{tab:physionet}, MDTIM achieves the lowest MAE across all settings, outperforming SAITS by a clear margin. Notably, FGTI exhibits the largest degradation here, even underperforming BRITS. This indicates that frequency-domain priors become unreliable for highly irregular signals where dominant spectral components are weak or unstable, whereas our ordinal-aware discrete formulation generalizes effectively to real-world sparse and irregular observations.

\subsection{Computational Efficiency and Model Scalability}
\label{sec:efficiency_analysis}
Diffusion-based methods often trade efficiency for accuracy. To examine this trade-off, we compare MDTIM against representative baselines at two model scales on the Energy dataset with uniform missing (Table~\ref{tab:performance_efficiency}).

\begin{table}[h!]
\caption{Imputation performance and efficiency on Energy (uniform missing). Inference time is measured on the full test set (MDTIM: $M{=}10$ noise samples; CSDI: $T{=}50$ diffusion steps).}
\label{tab:performance_efficiency}
\resizebox{1\columnwidth}{!}{%
\renewcommand{\arraystretch}{1.15}
\centering
\begin{tabular}{c|ccc|c|c|ccc|c|c}
\hline
\hline
\rowcolor{gray!20}
\textbf{Scale} & \multicolumn{5}{c|}{\textbf{Small}} & \multicolumn{5}{c}{\textbf{Large}} \\
\hline
\rowcolor{gray!20}
\textbf{Model} & \textbf{30\%} & \textbf{50\%} & \multicolumn{1}{c|}{\textbf{70\%}} & \textbf{Params (M)} & \textbf{Time (s)} & \textbf{30\%} & \textbf{50\%} & \multicolumn{1}{c|}{\textbf{70\%}} & \textbf{Params (M)} & \textbf{Time (s)} \\
\hline
BRITS         & 0.246 & 0.296 & 0.379 & 2.18  & 4.88   & 0.245 & 0.293 & 0.371 & 8.55  & 4.96 \\
SAITS         & 0.212 & 0.216 & 0.240 & 25.25 & 0.41   & 0.177 & 0.185 & 0.212 & 88.24 & 0.79 \\
CSDI          & 0.094 & 0.112 & 0.139 & 1.19  & 401.69 & 0.074 & 0.093 & 0.128 & 4.49  & 894.76 \\
MDTIM (Ours)  & \textbf{0.046} & \textbf{0.065} & \textbf{0.092} & 0.65  & 4.05   & \textbf{0.044} & \textbf{0.061} & \textbf{0.085} & 8.33 & 12.63 \\
\hline
\hline
\end{tabular}
}
\end{table}

MDTIM addresses the latency bottleneck of continuous diffusion: while CSDI requires up to 894s at the Large scale, the Small MDTIM completes inference in 4.05s while halving the MAE (0.074$\rightarrow$0.046). It is also notably parameter-efficient---the 0.65M Small variant surpasses the 88.24M Large SAITS, suggesting that the masked diffusion objective captures temporal dynamics more densely than scaling deterministic Transformers.

\subsection{Downstream Forecasting Task}

\begin{wraptable}{r}{0.4\columnwidth}
\vspace{-1.3em}
\centering
\caption{Forecasting MAE on Energy ($L{=}48$) using a fixed PatchTST predictor with imputed inputs from each model.}
\vspace{-0.5em}
\label{tab:forecasting}
\setlength{\tabcolsep}{3pt}
\renewcommand{\arraystretch}{1.05}
\resizebox{0.4\columnwidth}{!}{%
\begin{tabular}{l|cccc}
\hline
\hline
\rowcolor{gray!20}
\textbf{Energy} & \multicolumn{4}{c}{\textbf{Forecast Horizon}} \\
\hline
\rowcolor{gray!20}
\textbf{Imputer} & $H{=}4$ & $H{=}6$ & $H{=}8$ & $H{=}12$ \\
\hline
SAITS  & 0.187 & 0.207 & 0.224 & 0.254 \\
CSDI   & 0.171 & 0.190 & 0.208 & 0.237 \\
MDTIM (Ours)  & \textbf{0.138} & \textbf{0.153} & \textbf{0.170} & \textbf{0.200} \\
\hline
\hline
\end{tabular}}
\vspace{-0.5em}
\end{wraptable}

To assess whether the imputation accuracy of MDTIM translates to 
downstream tasks, we evaluate forecasting performance when the input 
window contains missing values. Each imputation model reconstructs the 
partially observed input, after which a fixed PatchTST~\cite{nie2023a} 
forecaster, pre-trained on clean sequences, predicts the subsequent $H$ 
steps. Since the forecaster is identical across all settings, differences 
in forecasting MAE directly reflect the quality of imputation.

As shown in Table~\ref{tab:forecasting}, MDTIM yields substantially 
lower forecasting error than both baselines across all horizons on 
Energy, reducing MAE by 19--26\% over SAITS and 16--19\% over CSDI. 
These results indicate that the imputation accuracy of MDTIM propagates 
to downstream forecasting, and that high-fidelity imputation is essential 
for tasks that rely on partially observed inputs.

\subsection{Ablation Studies}
\label{sec:ablation}

To rigorously evaluate the proposed framework, we conduct ablation studies on the mechanism by which missingness enters the model, the contribution of our expectation-based unmasking and soft labeling strategies, the spectral consistency term, and the sensitivity to the vocabulary size.

\paragraph{Missingness Mechanisms.}
\begin{wraptable}{r}{0.5\columnwidth}
\vspace{-1.0em}
\centering
\caption{MAE at 70\% missing when the \texttt{[MASK]} token is replaced by an attention mask that hides missing positions from observed ones.}
\vspace{-0.5em}
\label{tab:ablation_mask}
\resizebox{0.5\columnwidth}{!}{%
\begin{tabular}{l|cc|cc}
\hline
\hline
\rowcolor{gray!20}
& \multicolumn{2}{c|}{\textbf{Uniform}} & \multicolumn{2}{c}{\textbf{Geometric}} \\
\hline
\rowcolor{gray!20}
Dataset & Attn. mask & \texttt{[MASK]} & Attn. mask & \texttt{[MASK]} \\
\hline
ETTh    & 0.234  & \textbf{0.211}  & 0.225  & \textbf{0.205}  \\
Energy  & 0.087  & \textbf{0.085}  & 0.084  & \textbf{0.082}  \\
Weather & 0.047  & \textbf{0.044}  & 0.046  & \textbf{0.043}  \\
Sine    & 0.0038 & \textbf{0.0035} & 0.0036 & \textbf{0.0034} \\
\hline
\hline
\end{tabular}
}
\vspace{-1.0em}
\end{wraptable}
As argued in the introduction, what matters is how missingness enters the network, and we test this with two controls that keep the backbone and training identical.
The first replaces the \texttt{[MASK]} token by an attention mask: missing positions still read the observed ones, but they are removed from the keys of every attention block, so no observed position can see where the gaps are.
Table~\ref{tab:ablation_mask} shows that at 70\% missing this variant is worse on every dataset under both patterns, with the largest drop on ETTh, where the shape of a gap carries the most information about the values inside it.
The second control, MDTIM (Cont), keeps the masked diffusion training but operates on continuous values with a placeholder indicator, so the missing entries share the embedding space of the observations; Figure~\ref{fig:ablation} in Appendix~\ref{app:continuous_ablation} shows it trailing the discrete model at every missing rate on Energy.
Both mechanisms tell the model which entries are missing. The difference is that only the \texttt{[MASK]} token lets the rest of the sequence condition on that fact.

\paragraph{Impact of Unmasking and Labeling Strategies.}

\begin{table}[ht]
\vspace{-1.3em}
\centering
\caption{Ablation study on unmasking and labeling strategies. All results are reported as MAE averaged over 3 random mask seeds under the uniform missing scenario.}
\label{tab:ablation_components}
\resizebox{1\columnwidth}{!}{%
\begin{tabular}{cc|ccc|ccc|ccc}
\hline
\hline
\rowcolor{gray!20}
\multicolumn{2}{c|}{\textbf{Components}} & \multicolumn{3}{c|}{\textbf{ETTh}} & \multicolumn{3}{c|}{\textbf{Energy}} & \multicolumn{3}{c}{\textbf{Sine}} \\
\hline
\rowcolor{gray!20}
Exp. Unmasking & Soft-Label & 30\% & 50\% & 70\% & 30\% & 50\% & 70\% & 30\% & 50\% & 70\% \\
\hline
\xmark & \xmark         & 0.134 & 0.166 & 0.229 & 0.053 & 0.073 & 0.101 & 0.0071 & 0.0072 & 0.0075 \\
\xmark & \checkmark     & 0.131 & 0.162 & 0.222 & 0.056 & 0.079 & 0.115 & 0.0069 & 0.0070 & 0.0074 \\
\checkmark & \xmark     & 0.129 & 0.159 & 0.212 & 0.044 & \textbf{0.060} & 0.085 & 0.0026 & 0.0030 & 0.0036 \\
\checkmark & \checkmark & \textbf{0.127} & \textbf{0.157} & \textbf{0.211} & \textbf{0.044} & 0.061 & \textbf{0.085} & \textbf{0.0022} & \textbf{0.0026} & \textbf{0.0035} \\
\hline
\hline
\end{tabular}
}
\end{table}

Table~\ref{tab:ablation_components} summarizes the effect of our two proposed components: \textit{Expectation-based Unmasking} and \textit{Ordinal-Aware Soft Labeling}.
Expectation-based Unmasking consistently reduces error across all datasets, confirming that computing the expected value over discrete bins mitigates quantization error inherent in Argmax selection.
The contribution of Soft Labeling, in contrast, scales with the structural regularity of the data.
On the highly periodic Sine dataset, Soft Labeling yields a substantial additional gain on top of Expectation-based Unmasking, as ordinal-aware guidance aligns naturally with the smooth, predictable transitions of periodic signals.
A similar but milder gain is observed on ETTh, whose complex temporal dynamics still benefit from ordinal regularization under higher uncertainty.
On Energy, where the dynamics are less structured, Soft Labeling offers no further improvement beyond Expectation-based Unmasking.
These results indicate that Soft Labeling is most effective when the underlying signal exhibits clear ordinal or periodic structure that ordinal-aware guidance can exploit.

\paragraph{Impact of Spectral Consistency.}
\begin{wraptable}{r}{0.35\columnwidth}
\vspace{-1.0em}
\centering
\caption{MAE on ETTh with and without the spectral term ($\lambda$) across window lengths $L$, uniform missing.}
\vspace{-0.5em}
\label{tab:ablation_spectral}
\resizebox{0.35\columnwidth}{!}{%
\begin{tabular}{cc|ccc}
\hline
\hline
\rowcolor{gray!20}
\multicolumn{2}{c|}{\textbf{ETTh}} & \multicolumn{3}{c}{\textbf{Length ($L$)}} \\
\hline
\rowcolor{gray!20}
Rate & $\lambda$ & 48 & 96 & 192 \\
\hline
\multirow{2}{*}{30\%} & 0 & \textbf{0.127} & 0.125 & 0.126 \\
                      & 1 & \textbf{0.127} & \textbf{0.124} & \textbf{0.123} \\
\hline
\multirow{2}{*}{50\%} & 0 & \textbf{0.157} & \textbf{0.152} & 0.151 \\
                      & 1 & \textbf{0.157} & \textbf{0.152} & \textbf{0.148} \\
\hline
\multirow{2}{*}{70\%} & 0 & \textbf{0.208} & \textbf{0.196} & 0.191 \\
                      & 1 & 0.211 & 0.199 & \textbf{0.190} \\
\hline
\hline
\end{tabular}
}
\vspace{-1.0em}
\end{wraptable}
Table~\ref{tab:ablation_spectral} removes the spectral term ($\lambda{=}0$) with all other components in place, on ETTh, the most broadband of our benchmarks.
At $L{=}48$ the two variants are tied at 30\% and 50\% and $\lambda{=}0$ is marginally ahead at 70\%, so on a short window the term has little to work with.
This is a matter of frequency resolution rather than a harmful objective: the resolution of the FFT is set by the window length, and as $L$ grows to 192 the spectral term pulls ahead at every missing rate.
The same reasoning governs PhysioNet, where the observed entries are too sparse to form a reliable spectrum and we set $\lambda{=}0$.
Even without the spectral term, MDTIM remains ahead of every baseline in Table~\ref{tab:main_results1}, so the reported gains rest on the discrete formulation rather than on the auxiliary loss.

\paragraph{Sensitivity to Vocabulary Size ($K$).}
\begin{wraptable}{r}{0.45\columnwidth}
\vspace{-1.0em}
\centering
\caption{MAE on ETTh with varying vocabulary size ($K$).}
\vspace{-0.5em}
\label{tab:ablation_bins}
\resizebox{0.45\columnwidth}{!}{%
\begin{tabular}{c|ccc|ccc}
\hline
\hline
\rowcolor{gray!20}
\multicolumn{1}{c|}{\textbf{ETTh}} & \multicolumn{3}{c|}{\textbf{Uniform}} & \multicolumn{3}{c}{\textbf{Geometric}} \\
\hline
\rowcolor{gray!20}
Bins & 30\% & 50\% & 70\% & 30\% & 50\% & 70\% \\
\hline
20 & 0.138 & 0.166 & 0.217 & 0.157 & 0.181 & 0.211 \\
40 & \textbf{0.127} & \textbf{0.157} & \textbf{0.211} & \textbf{0.146} & \textbf{0.173} & \textbf{0.205} \\
60 & 0.129 & 0.160 & 0.214 & 0.148 & 0.175 & 0.208 \\
\hline
\hline
\end{tabular}
}
\vspace{-1.0em}
\end{wraptable}
We further investigate the impact of the vocabulary size $K$ on imputation accuracy (Table~\ref{tab:ablation_bins}), which exhibits a trade-off between discretization resolution and classification complexity.
A small vocabulary ($K{=}20$) suffers from high quantization error, failing to capture fine-grained fluctuations.
Conversely, $K{=}60$ slightly degrades performance, as a larger vocabulary increases the difficulty of the discrete classification task.
$K{=}40$ achieves the best balance, providing sufficient resolution to approximate the continuous signal while remaining tractable to optimize.

\section{Conclusion and Limitations}

We proposed MDTIM, a framework that adapts the masked diffusion training 
paradigm to continuous time series imputation. By introducing Stochastic 
Discretization and Ordinal-Aware Soft Labeling, MDTIM bridges the gap 
between continuous dynamics and categorical tokenization, while the 
Expectation-based Unmasking strategy enables precise continuous 
reconstruction. Extensive experiments demonstrate that MDTIM consistently 
outperforms both deterministic and generative baselines in reconstruction 
accuracy and robustness across diverse missing scenarios, while requiring 
substantially less inference time than continuous diffusion baselines.

% \paragraph{Limitations and Future Work.} 
% While MDTIM demonstrates strong performance across diverse missing 
% scenarios, several aspects warrant further investigation. First, the 
% optimal vocabulary size $K$ may vary across datasets depending on the 
% range and granularity of the underlying signal; a 
% data-adaptive discretization scheme that automatically selects the 
% resolution would further improve generalization. Second, our spectral 
% consistency regularization assumes the presence of stable frequency 
% components and is therefore disabled on benchmarks dominated by 
% irregular sampling such as PhysioNet. Designing an 
% adaptive frequency objective that modulates its influence based on 
% local signal regularity is a promising direction. Finally, our 
% evaluation focuses on standard imputation benchmarks of moderate 
% length; extending MDTIM to extremely long sequences and to multi-scale 
% or non-stationary domains such as financial tick data remains for 
% future work.

\paragraph{Limitations and Future Work.} While MDTIM demonstrates strong performance across diverse missing scenarios, several aspects warrant further investigation. First, Stochastic Discretization is unbiased in expectation but, like any quantization, lossy for an individual sample, and a data-adaptive discretization scheme that automatically selects the resolution would further improve generalization. Second, our spectral consistency regularization assumes the presence of stable frequency components and is therefore disabled on benchmarks dominated by irregular sampling such as PhysioNet. Designing an adaptive frequency objective that modulates its influence based on local signal regularity is a promising direction. Third, the per-channel embedding tables cost $O(CKD)$ parameters, which grows linearly with the number of variates; a channel-shared alternative is examined in Appendix~\ref{app:shared}. Finally, our evaluation focuses on standard imputation benchmarks of moderate length; extending MDTIM to extremely long sequences and to multi-scale or non-stationary domains such as financial tick data remains for future work.

\section*{Acknowledgments}
This work was supported by Basic Science Research Program through the National Research Foundation of Korea (NRF) funded by the Ministry of Education (No. RS-2025-25420430). This work was also partly supported by the National Research Foundation of Korea (NRF) grant funded by the Korean government (MSIT) (No. RS-2024-00338859), by the Technology Innovation Program (No. RS-2026-25544725) funded by the Ministry of Trade, Industry and Resources (MOTIR, Korea), and by the Ministry of Trade, Industry and Resources (MOTIR), Republic of Korea, in 2026, under the project ``Development of Core Foundation Model and AI Agent Technology for Thermal-Process Manufacturing (TP-MFM)'' (No. TPAI2616). All four grants contributed equally to this work.

\bibliography{ref}

@article{yang2024frequency,
  title={Frequency-aware generative models for multivariate time series imputation},
  author={Yang, Xinyu and Sun, Yu and Yuan, Xiaojie and Chen, Xinyang},
  journal={Advances in Neural Information Processing Systems},
  volume={37},
  pages={52595--52623},
  year={2024}
}

@inproceedings{nie2024imputeformer,
  title={ImputeFormer: Low rankness-induced transformers for generalizable spatiotemporal imputation},
  author={Nie, Tong and Qin, Guoyang and Ma, Wei and Mei, Yuewen and Sun, Jian},
  booktitle={Proceedings of the 30th ACM SIGKDD conference on knowledge discovery and data mining},
  pages={2260--2271},
  year={2024}
}

@article{che2018recurrent,
  title={Recurrent neural networks for multivariate time series with missing values},
  author={Che, Zhengping and Purushotham, Sanjay and Cho, Kyunghyun and Sontag, David and Liu, Yan},
  journal={Scientific reports},
  volume={8},
  number={1},
  pages={6085},
  year={2018},
  publisher={Nature Publishing Group UK London}
}

@article{cao2018brits,
  title={Brits: Bidirectional recurrent imputation for time series},
  author={Cao, Wei and Wang, Dong and Li, Jian and Zhou, Hao and Li, Lei and Li, Yitan},
  journal={Advances in neural information processing systems},
  volume={31},
  year={2018}
}

@article{du2023saits,
  title={Saits: Self-attention-based imputation for time series},
  author={Du, Wenjie and C{\^o}t{\'e}, David and Liu, Yan},
  journal={Expert Systems with Applications},
  volume={219},
  pages={119619},
  year={2023},
  publisher={Elsevier}
}

@inproceedings{fortuin2020gp,
  title={Gp-vae: Deep probabilistic time series imputation},
  author={Fortuin, Vincent and Baranchuk, Dmitry and R{\"a}tsch, Gunnar and Mandt, Stephan},
  booktitle={International conference on artificial intelligence and statistics},
  pages={1651--1661},
  year={2020},
  organization={PMLR}
}

@article{tashiro2021csdi,
  title={Csdi: Conditional score-based diffusion models for probabilistic time series imputation},
  author={Tashiro, Yusuke and Song, Jiaming and Song, Yang and Ermon, Stefano},
  journal={Advances in neural information processing systems},
  volume={34},
  pages={24804--24816},
  year={2021}
}

@inproceedings{silva2012predicting,
  title={Predicting in-hospital mortality of icu patients: The physionet/computing in cardiology challenge 2012},
  author={Silva, Ikaro and Moody, George and Scott, Daniel J and Celi, Leo A and Mark, Roger G},
  booktitle={2012 computing in cardiology},
  pages={245--248},
  year={2012},
  organization={IEEE}
}

@article{austin2021structured,
  title={Structured denoising diffusion models in discrete state-spaces},
  author={Austin, Jacob and Johnson, Daniel D and Ho, Jonathan and Tarlow, Daniel and Van Den Berg, Rianne},
  journal={Advances in neural information processing systems},
  volume={34},
  pages={17981--17993},
  year={2021}
}

@article{sahoo2024simple,
  title={Simple and effective masked diffusion language models},
  author={Sahoo, Subham and Arriola, Marianne and Schiff, Yair and Gokaslan, Aaron and Marroquin, Edgar and Chiu, Justin and Rush, Alexander and Kuleshov, Volodymyr},
  journal={Advances in Neural Information Processing Systems},
  volume={37},
  pages={130136--130184},
  year={2024}
}

@inproceedings{kim2025train,
  title={Train for the Worst, Plan for the Best: Understanding Token Ordering in Masked Diffusions},
  author={Kim, Jaeyeon and Shah, Kulin and Kontonis, Vasilis and Kakade, Sham M and Chen, Sitan},
  booktitle={Forty-second International Conference on Machine Learning},
  year={2025}
}

@inproceedings{koh2025conditional,
  title={Conditional [MASK] Discrete Diffusion Language Model},
  author={Koh, Hyukhun and Jhang, Minha and Kim, Dohyung and Lee, Sangmook and Jung, Kyomin},
  booktitle={Proceedings of the 2025 Conference on Empirical Methods in Natural Language Processing},
  pages={8910--8934},
  year={2025}
}

@article{rout2025anchored,
  title={Anchored Diffusion Language Model},
  author={Rout, Litu and Caramanis, Constantine and Shakkottai, Sanjay},
  journal={Advances in Neural Information Processing Systems},
  year={2025}
}

@article{hoogeboom2021argmax,
  title={Argmax flows and multinomial diffusion: Learning categorical distributions},
  author={Hoogeboom, Emiel and Nielsen, Didrik and Jaini, Priyank and Forr, Patrick and Welling, Max},
  journal={Advances in neural information processing systems},
  volume={34},
  pages={12454--12465},
  year={2021}
}

@article{vaswani2017attention,
  title={Attention is all you need},
  author={Vaswani, Ashish and Shazeer, Noam and Parmar, Niki and Uszkoreit, Jakob and Jones, Llion and Gomez, Aidan N and Kaiser, {\L}ukasz and Polosukhin, Illia},
  journal={Advances in neural information processing systems},
  volume={30},
  year={2017}
}

@article{yoon2018estimating,
  title={Estimating missing data in temporal data streams using multi-directional recurrent neural networks},
  author={Yoon, Jinsung and Zame, William R and Van Der Schaar, Mihaela},
  journal={IEEE Transactions on Biomedical Engineering},
  volume={66},
  number={5},
  pages={1477--1490},
  year={2018},
  publisher={IEEE}
}

@inproceedings{zhou2021informer,
  title={Informer: Beyond efficient transformer for long sequence time-series forecasting},
  author={Zhou, Haoyi and Zhang, Shanghang and Peng, Jieqi and Zhang, Shuai and Li, Jianxin and Xiong, Hui and Zhang, Wancai},
  booktitle={Proceedings of the AAAI conference on artificial intelligence},
  volume={35},
  number={12},
  pages={11106--11115},
  year={2021}
}

@article{wu2021autoformer,
  title={Autoformer: Decomposition transformers with auto-correlation for long-term series forecasting},
  author={Wu, Haixu and Xu, Jiehui and Wang, Jianmin and Long, Mingsheng},
  journal={Advances in neural information processing systems},
  volume={34},
  pages={22419--22430},
  year={2021}
}

@inproceedings{zeng2023transformers,
  title={Are transformers effective for time series forecasting?},
  author={Zeng, Ailing and Chen, Muxi and Zhang, Lei and Xu, Qiang},
  booktitle={Proceedings of the AAAI conference on artificial intelligence},
  volume={37},
  number={9},
  pages={11121--11128},
  year={2023}
}

@article{zhou2022film,
  title={Film: Frequency improved legendre memory model for long-term time series forecasting},
  author={Zhou, Tian and Ma, Ziqing and Wen, Qingsong and Sun, Liang and Yao, Tao and Yin, Wotao and Jin, Rong and others},
  journal={Advances in neural information processing systems},
  volume={35},
  pages={12677--12690},
  year={2022}
}

@article{yi2023frequency,
  title={Frequency-domain MLPs are more effective learners in time series forecasting},
  author={Yi, Kun and Zhang, Qi and Fan, Wei and Wang, Shoujin and Wang, Pengyang and He, Hui and An, Ning and Lian, Defu and Cao, Longbing and Niu, Zhendong},
  journal={Advances in Neural Information Processing Systems},
  volume={36},
  pages={76656--76679},
  year={2023}
}

@inproceedings{
wu2023timesnet,
title={TimesNet: Temporal 2D-Variation Modeling for General Time Series Analysis},
author={Haixu Wu and Tengge Hu and Yong Liu and Hang Zhou and Jianmin Wang and Mingsheng Long},
booktitle={The Eleventh International Conference on Learning Representations },
year={2023},
url={https://openreview.net/forum?id=ju_Uqw384Oq}
}

@article{liu2022scinet,
  title={Scinet: Time series modeling and forecasting with sample convolution and interaction},
  author={Liu, Minhao and Zeng, Ailing and Chen, Muxi and Xu, Zhijian and Lai, Qiuxia and Ma, Lingna and Xu, Qiang},
  journal={Advances in Neural Information Processing Systems},
  volume={35},
  pages={5816--5828},
  year={2022}
}

@inproceedings{miao2021generative,
  title={Generative semi-supervised learning for multivariate time series imputation},
  author={Miao, Xiaoye and Wu, Yangyang and Wang, Jun and Gao, Yunjun and Mao, Xudong and Yin, Jianwei},
  booktitle={Proceedings of the AAAI conference on artificial intelligence},
  volume={35},
  number={10},
  pages={8983--8991},
  year={2021}
}

@inproceedings{
yuan2024diffusionts,
title={Diffusion-{TS}: Interpretable Diffusion for General Time Series Generation},
author={Xinyu Yuan and Yan Qiao},
booktitle={The Twelfth International Conference on Learning Representations},
year={2024},
url={https://openreview.net/forum?id=4h1apFjO99}
}

@inproceedings{
nie2023a,
title={A Time Series is Worth 64 Words:  Long-term Forecasting with Transformers},
author={Yuqi Nie and Nam H Nguyen and Phanwadee Sinthong and Jayant Kalagnanam},
booktitle={The Eleventh International Conference on Learning Representations },
year={2023},
url={https://openreview.net/forum?id=Jbdc0vTOcol}
}

@article{du2024tsi,
  title={Tsi-bench: Benchmarking time series imputation},
  author={Du, Wenjie and Wang, Jun and Qian, Linglong and Yang, Yiyuan and Ibrahim, Zina and Liu, Fanxing and Wang, Zepu and Liu, Haoxin and Zhao, Zhiyuan and Zhou, Yingjie and others},
  journal={arXiv preprint arXiv:2406.12747},
  year={2024}
}

@inproceedings{peebles2023scalable,
  title={Scalable diffusion models with transformers},
  author={Peebles, William and Xie, Saining},
  booktitle={Proceedings of the IEEE/CVF international conference on computer vision},
  pages={4195--4205},
  year={2023}
}

@article{su2024roformer,
  title={Roformer: Enhanced transformer with rotary position embedding},
  author={Su, Jianlin and Ahmed, Murtadha and Lu, Yu and Pan, Shengfeng and Bo, Wen and Liu, Yunfeng},
  journal={Neurocomputing},
  volume={568},
  pages={127063},
  year={2024},
  publisher={Elsevier}
}

@inproceedings{10.24963/ijcai.2025/1187,
author = {Wang, Jun and Du, Wenjie and Yang, Yiyuan and Qian, Linglong and Cao, Wei and Zhang, Keli and Wang, Wenjia and Liang, Yuxuan and Wen, Qingsong},
title = {Deep learning for multivariate time series imputation: a survey},
year = {2025},
isbn = {978-1-956792-06-5},
url = {https://doi.org/10.24963/ijcai.2025/1187},
doi = {10.24963/ijcai.2025/1187},
booktitle = {Proceedings of the Thirty-Fourth International Joint Conference on Artificial Intelligence},
articleno = {1187},
numpages = {9},
location = {Montreal, Canada},
series = {IJCAI '25}
}

@inproceedings{
fadlon2025a,
title={A Diffusion Model for Regular Time Series Generation from Irregular Data with Completion and Masking},
author={Gal Fadlon and Idan Arbiv and Nimrod Berman and Omri Azencot},
booktitle={The Thirty-ninth Annual Conference on Neural Information Processing Systems},
year={2025},
url={https://openreview.net/forum?id=M9JmlA6Cgf}
}

@inproceedings{kim2023probabilistic,
  title={Probabilistic imputation for time-series classification with missing data},
  author={Kim, SeungHyun and Kim, Hyunsu and Yun, Eunggu and Lee, Hwangrae and Lee, Jaehun and Lee, Juho},
  booktitle={International Conference on Machine Learning},
  pages={16654--16667},
  year={2023},
  organization={PMLR}
}

@article{matheson1976scoring,
  title={Scoring rules for continuous probability distributions},
  author={Matheson, James E and Winkler, Robert L},
  journal={Management science},
  volume={22},
  number={10},
  pages={1087--1096},
  year={1976},
  publisher={INFORMS}
}

@inproceedings{sohl2015deep,
  title={Deep unsupervised learning using nonequilibrium thermodynamics},
  author={Sohl-Dickstein, Jascha and Weiss, Eric and Maheswaranathan, Niru and Ganguli, Surya},
  booktitle={International conference on machine learning},
  pages={2256--2265},
  year={2015},
  organization={pmlr}
}

@article{ho2020denoising,
  title={Denoising diffusion probabilistic models},
  author={Ho, Jonathan and Jain, Ajay and Abbeel, Pieter},
  journal={Advances in neural information processing systems},
  volume={33},
  pages={6840--6851},
  year={2020}
}

@article{song2020score,
  title={Score-based generative modeling through stochastic differential equations},
  author={Song, Yang and Sohl-Dickstein, Jascha and Kingma, Diederik P and Kumar, Abhishek and Ermon, Stefano and Poole, Ben},
  journal={arXiv preprint arXiv:2011.13456},
  year={2020}
}

@article{dhariwal2021diffusion,
  title={Diffusion models beat gans on image synthesis},
  author={Dhariwal, Prafulla and Nichol, Alexander},
  journal={Advances in neural information processing systems},
  volume={34},
  pages={8780--8794},
  year={2021}
}

@article{ho2022classifier,
  title={Classifier-free diffusion guidance},
  author={Ho, Jonathan and Salimans, Tim},
  journal={arXiv preprint arXiv:2207.12598},
  year={2022}
}

@article{alcaraz2022diffusion,
  title={Diffusion-based time series imputation and forecasting with structured state space models},
  author={Alcaraz, Juan Miguel Lopez and Strodthoff, Nils},
  journal={arXiv preprint arXiv:2208.09399},
  year={2022}
}

@inproceedings{rombach2022high,
  title={High-resolution image synthesis with latent diffusion models},
  author={Rombach, Robin and Blattmann, Andreas and Lorenz, Dominik and Esser, Patrick and Ommer, Bj{\"o}rn},
  booktitle={Proceedings of the IEEE/CVF conference on computer vision and pattern recognition},
  pages={10684--10695},
  year={2022}
}

@article{karras2022elucidating,
  title={Elucidating the design space of diffusion-based generative models},
  author={Karras, Tero and Aittala, Miika and Aila, Timo and Laine, Samuli},
  journal={Advances in neural information processing systems},
  volume={35},
  pages={26565--26577},
  year={2022}
}

@article{kong2020diffwave,
  title={Diffwave: A versatile diffusion model for audio synthesis},
  author={Kong, Zhifeng and Ping, Wei and Huang, Jiaji and Zhao, Kexin and Catanzaro, Bryan},
  journal={arXiv preprint arXiv:2009.09761},
  year={2020}
}

@article{ho2022video,
  title={Video diffusion models},
  author={Ho, Jonathan and Salimans, Tim and Gritsenko, Alexey and Chan, William and Norouzi, Mohammad and Fleet, David J},
  journal={Advances in neural information processing systems},
  volume={35},
  pages={8633--8646},
  year={2022}
}

@article{lopez2023diffusion,
  title={Diffusion-based time series imputation and forecasting with structured atate apace models},
  author={Lopez Alcaraz, Juan Miguel and Strodthoff, Nils},
  journal={Transactions on machine learning research},
  pages={1--36},
  year={2023}
}
\bibliographystyle{plainnat}
%%%%%%%%%%%%%%%%%%%%%%%%%%%%%%%%%%%%%%%%%%%%%%%%%%%%%%%%%%%%
\newpage
\appendix

\section{Preliminaries on Diffusion Models}
\subsection{Diffusion Models}
\label{app:dm_bg}

Diffusion models~\citep{sohl2015deep,ho2020denoising,song2020score} are a class of latent-variable generative models that learn to invert a fixed noise-injection process.
Since their introduction, they have become the dominant paradigm in many continuous-data generative tasks, achieving state-of-the-art results in image~\citep{dhariwal2021diffusion,
rombach2022high,karras2022elucidating},
audio~\citep{kong2020diffwave}, and video~\citep{ho2022video} synthesis, and have also been adopted as strong probabilistic models for time series forecasting and imputation~\citep{tashiro2021csdi,alcaraz2022diffusion}.

\paragraph{Notation.}
Let $x_0\in\mathbb{R}^d$ denote a clean data sample drawn from the data distribution $q(x_0)$, and let $x_t\in\mathbb{R}^d$ for $t\in\{1,\dots,T\}$ denote its noisy latents along the forward chain.
A noise schedule $\{\beta_t\}_{t=1}^{T}\subset(0,1)$ controls the amount of corruption at each step, with the per-step retention factor $\alpha_t=1-\beta_t$ and the cumulative retention factor $\bar{\alpha}_t=\prod_{s\le t}\alpha_s$; by construction
$\bar{\alpha}_0 = 1$ and $\bar{\alpha}_T \approx 0$.

\subsubsection{Diffusion Framework}
\label{app:dm_model}

\paragraph{Forward (Gaussian corruption).}

The forward process gradually corrupts a data sample into pure Gaussian noise over $T$ steps:
\begin{equation}
q(x_t\mid x_{t-1}) = \mathcal{N} (x_t;\sqrt{1-\beta_t} x_{t-1}, \beta_t I).
\label{eq:app_forward}
\end{equation}
A useful property of this Markov chain is that its $t$-step marginal admits a closed form:
\begin{equation}
q(x_t \vert x_0) = \mathcal{N} (x_t;\sqrt{\bar{\alpha}_t} x_0, (1-\bar{\alpha}_t)I),
\
x_t=\sqrt{\bar{\alpha}_t} x_0+\sqrt{1-\bar{\alpha}_t} \epsilon, \ \epsilon\sim\mathcal{N}(0,I).
\label{eq:app_forward_closed_form}
\end{equation}
That is, $x_t$ at any timestep $t$ can be sampled \emph{directly} from
$x_0$ without simulating the full chain, which makes per-step training tractable.
As $t\to T$, the schedule is chosen so that $\bar{\alpha}_T \approx 0$ and $x_T$ becomes approximately isotropic Gaussian, independent of $x_0$.

\paragraph{Reverse (learned denoising).}
The reverse process incrementally denoises noise back into the data manifold.
While the marginal $q(x_{t-1}\vert x_t)$
is intractable, the \emph{conditional} posterior $q(x_{t-1}\vert x_t,x_0)$ is Gaussian with closed-form mean and variance depending only on $\beta_t$ and $\bar{\alpha}_t$~\citep{ho2020denoising}.
A neural network parameterizes the reverse kernel
\begin{equation}
p_\theta(x_{t-1}\vert x_t) = \mathcal{N} (x_{t-1}; \mu_\theta(x_t,t),  \Sigma_\theta(x_t,t)),
\label{eq:app_reverse}
\end{equation}
where the mean is typically expressed through a noise predictor
$\epsilon_\theta(x_t,t)$:
\begin{equation}
\mu_\theta(x_t,t)
=\frac{1}{\sqrt{\alpha_t}}\Big(x_t - \frac{\beta_t}{\sqrt{1-\bar{\alpha}_t}} \epsilon_\theta(x_t,t)\Big).
\label{eq:app_mu_theta}
\end{equation}

\subsubsection{Conditional Diffusion}
\label{app:dm_cond}

The diffusion framework introduced in Appendix~\ref{app:dm_bg} defines an unconditional generative model.
Many practical applications, including the time series imputation setting of this paper, instead require sampling from a conditional distribution $p(x_0\mid y)$.

Let $\mathcal{O}\subseteq\{1,\dots,d\}$ index observed coordinates and $\mathcal{M}=\bar{\mathcal{O}}$ the missing ones.
The goal is to sample
\begin{equation}
x_0^{\mathcal{M}}\sim p(x_0^{\mathcal{M}}\mid x_0^{\mathcal{O}}).
\end{equation}

While external-conditioning paradigms such as classifier 
guidance~\citep{dhariwal2021diffusion} and classifier-free guidance~\citep{ho2022classifier} are effective 
when $y$ is an external attribute such as a class label or text prompt, they do not apply directly to time series imputation, where the conditioning signal is a subset of the data itself.
Diffusion-based imputation methods therefore commonly adopt the following setting~\citep{tashiro2021csdi, alcaraz2022diffusion}.

The forward process applies Gaussian corruption only to missing coordinates while preserving observed ones throughout all steps,
\begin{equation}
q(x_t^{\mathcal{M}}\vert x_0^{\mathcal{M}}) 
= \mathcal{N}(x_t^{\mathcal{M}}; \sqrt{\bar{\alpha}_t}\,x_0^{\mathcal{M}}, 
(1-\bar{\alpha}_t) I), 
\quad 
x_t^{\mathcal{O}} = x_0^{\mathcal{O}} \ \forall t.
\label{eq:app_cond_forward}
\end{equation}
The noise predictor receives the partially-noisy tensor 
$\tilde{x}_t$ (noisy on $\mathcal{M}$, clean on $\mathcal{O}$) together with the missingness mask $\mathcal{M}$ and the timestep $t$, $\epsilon_\theta(\tilde{x}_t,\mathcal{M},t)$, giving the network direct access to which coordinates are observed 
and what their values are.

\subsubsection{Training objective}
\label{app:dm_training}

Training maximizes a variational lower bound (ELBO) on $\log p_\theta(x_0)$.
The bound decomposes into a sum of KL divergences between the tractable forward posterior $q(x_{t-1}\mid x_t,x_0)$ and the learned reverse kernel $p_\theta(x_{t-1}\mid x_t)$ at each step~\citep{ho2020denoising}.
Reparameterizing the network as a noise predictor 
$\epsilon_\theta$ and substituting the closed-form expression for $x_t$ in Equation~\eqref{eq:app_forward_closed_form}, 
the per-step KL reduces (up to a $t$-dependent reweighting) to the simple denoising loss of~\citet{ho2020denoising}:
\begin{equation}
\mathcal{L}_{\mathrm{simple}}(\theta)
=
\mathbb{E}_{t \sim\mathcal{U}\{1,\dots,T\} ,x_0,\epsilon\sim\mathcal{N}(0,I)}
[\Vert \epsilon-\epsilon_\theta (\sqrt{\bar{\alpha}_t}x_0+\sqrt{1-\bar{\alpha}_t}\epsilon, t)\Vert_2^{2}].
\label{eq:app_noise_pred}
\end{equation}

For the conditional setting in Appendix~\ref{app:dm_cond}, the noise predictor 
$\epsilon_\theta(\tilde{x}_t,\mathcal{M},t)$ takes the 
partially noisy tensor and the missingness mask as input, 
and the loss is restricted to the missing coordinates~\citep{tashiro2021csdi,alcaraz2022diffusion}:
\begin{equation}
\mathcal{L}_{\mathrm{imp}}(\theta)
=
\mathbb{E}_{t \sim\mathcal{U}\{1,\dots,T\} ,x_0,\epsilon\sim\mathcal{N}(0,I),\mathcal{M}}
\left[\,\Vert \epsilon^{\mathcal{M}} 
- \epsilon_\theta^{\mathcal{M}}(\tilde{x}_t,\mathcal{M},t)
\Vert_2^{2}\,\right],
\label{eq:app_imp_loss}
\end{equation}
where the missingness mask $\mathcal{M}$ is sampled from a distribution of patterns during training, encouraging the model to generalize across missingness configurations.

\subsection{Masked Diffusion Language Models}
\label{app:mdlm_bg}

Masked Diffusion Language Models (MDLM)~\citep{sahoo2024simple} adapt the diffusion framework of Appendix~\ref{app:dm_bg} to discrete sequences by replacing Gaussian corruption with an absorbing-state Markov chain~\citep{austin2021structured}.
The role of pure noise $\mathcal{N}(0,I)$ in continuous diffusion is played here by a single distinguished symbol \texttt{[MASK]}: the forward process gradually replaces tokens with \texttt{[MASK]}, the prior at $t=1$ is the all-\texttt{[MASK]} sequence, and the reverse process iteratively \emph{unmasks} positions back to data tokens.

\paragraph{Notation.}
Let $t(i)=\frac{i}{T}$ and $s(i)=\frac{i-1}{T}$ for $i\in\{1,\dots,T\}$.
Tokens take values in $\{0,1,\dots,K\}$ where $0$ denotes \texttt{[MASK]}.
We represent tokens as one-hot vectors in $\mathcal{V}\coloneqq\{e_0,\dots,e_K\}\subset\{0,1\}^{K+1}$ and denote the \texttt{[MASK]} one-hot by $m\coloneqq e_0$.
For a length-$L$ sequence $x=(x^l)_{l=1}^L$, let $z_t=(z_t^l)_{l=1}^L$ be the latent at time $t$.
We write $\mathrm{Cat}(\cdot;\pi)$ for a categorical distribution and $\langle\cdot,\cdot\rangle$ for the dot product.

\subsubsection{MDLM Framework}
\label{app:mdlm_model}

\paragraph{Forward (absorbing mask corruption).}

The forward process factorizes across token positions and replaces each token by \texttt{[MASK]} with probability $1-\alpha_t$:
\begin{equation}
q(z_t\mid x)=\prod_{l=1}^L \mathrm{Cat}(z_t^l; \alpha_t x^l + (1-\alpha_t)m),
\label{eq:app_mdlm_forward}
\end{equation}
with $\alpha_t \to 1$ as $t\to 0$ and $\alpha_t \to 0$ as $t\to 1$.
We parameterize the noise schedule via $\alpha_t = e^{-\sigma(t)}$.
Our experiments use the \textit{log-linear} schedule defined by $\sigma(t) = -\log(1-t)$, which yields a linear decay $\alpha_t = 1-t$.

The defining property of this process is that \texttt{[MASK]} is an \emph{absorbing} state: once $z_t^l=m$, it remains masked for all later times.
Consequently, $q(z_t\vert x)$ depends only on which positions have been masked and the reverse process is naturally structured around iterative
\emph{unmasking}.

\paragraph{Reverse (learned denoising kernel).}
MDLM parameterizes the reverse kernel $p_\theta(z_s\mid z_t)=\prod_{l=1}^L p_\theta(z_s^l\mid z_t)$ via a predicted mixture token $x_\theta^l(z_t)$ over the $K$ non-mask vocabulary entries.
Two structural constraints mirror the
absorbing forward process:
(i) \emph{zero-masking}, $\langle x_\theta^l(z_t), m\rangle = 0$, so the
predictor never re-introduces \texttt{[MASK]} and (ii) \emph{carry-over}, $z_t^l\neq m \Rightarrow z_s^l=z_t^l$, so
already-unmasked tokens are preserved. As a result, the reverse transition is
\begin{equation}
p_\theta(z_s^l\mid z_t) = 
\begin{cases}
\mathrm{Cat}(z_s^l; z_t^l), & z_t^l\neq m, \\ \mathrm{Cat}(z_s^l; \frac{\alpha_s-\alpha_t}{1-\alpha_t}x_\theta^l(z_t) + \frac{1-\alpha_s}{1-\alpha_t}m), & z_t^l=m.
\end{cases}
\label{eq:app_mdlm_reverse}
\end{equation}
For masked positions, the update is a mixture: with probability $\frac{\alpha_s-\alpha_t}{1-\alpha_t}$ the token is unmasked according
to $x_\theta^l$, and otherwise it remains \texttt{[MASK]}.

\subsubsection{Conditional Masking}
\label{app:mdlm_cond}

For text infilling, observed tokens can be treated as anchors and excluded from corruption.
Let $\mathcal{A}\subseteq\{1,\dots,L\}$ be the anchored index set and $\bar{\mathcal{A}}$ its complement.
Anchored positions are preserved exactly throughout the forward process while non-anchored positions undergo standard absorbing mask corruption:
\begin{equation}
q(z_t \mid x)
=
\prod_{l\in\mathcal{A}} \mathbb{I}[z_t^l = x^l]
\prod_{l\in\bar{\mathcal{A}}}
\mathrm{Cat}\!\left(z_t^l;\; \alpha_t x^l + (1-\alpha_t)m\right),
\label{eq:app_anchored_forward}
\end{equation}
so that $z_t^l=x^l$ for all $t$ and all $l\in\mathcal{A}$ (hard preservation).

The reverse kernel mirrors this structure.
Anchored positions are copied from $z_t$ unchanged, and only non-anchored masked positions are denoised via the standard reverse kernel in Equation~\eqref{eq:app_mdlm_reverse}.
This guarantees that observed tokens remain unchanged throughout generation while the model fills in the unobserved ones.

This anchored formulation directly matches the imputation setting in Appendix~\ref{app:dm_cond}.
The anchored set $\mathcal{A}$ corresponds to the observed coordinates $\mathcal{O}$, and the complement $\bar{\mathcal{A}}$ corresponds to the missing coordinates $\mathcal{M}$.

\subsubsection{Training Objective}
\label{app:mdlm_training}

MDLM trains the denoising kernel by minimizing a negative ELBO (NELBO).
Owing to the absorbing structure of the forward process, each per-step KL between the forward posterior and the learned reverse kernel reduces in closed form to a weighted cross-entropy over currently masked positions, $w_{t} \sum_{l \in \mathcal{M}_t} \mathrm{CE}(x_\theta^l(z_t), x^l)$ where $w_{t} = \tfrac{\alpha_t - \alpha_s}{1 - \alpha_t}$ and $\mathcal{M}_t = 
\{l \mid z_t^l = m\}$ is the set of currently masked positions.
The denoiser $x_\theta^l(z_t)$ predicts the clean token at position $l$ directly, paralleling the prediction of clean signal $x_0$ in continuous diffusion.
Taking $T\to\infty$, the discrete-time sum converges to a 
continuous-time objective that samples $t\sim\mathcal{U}(0,1)$ and uses the weight $w(t) = \frac{\alpha_t'}{1-\alpha_t}$:
\begin{equation}
\mathcal{L}_{\mathrm{CT}}(x;\theta)
=
\mathbb{E}_{t}
\left[
\frac{\alpha_t'}{1-\alpha_t}
\sum_{l\in\mathcal{M}_t}
\mathrm{CE}\!\left(x_\theta^l(z_t), x^l\right)
\right],
\label{eq:app_mdlm_ct}
\end{equation}
which integrates the per-step loss over a continuous time grid and is the form actually used in our experiments.

With anchors (Section~\ref{app:mdlm_cond}), observed positions are excluded from the forward process and therefore never appear masked at any time $t$.
The training loss is consequently computed only on non-anchor masked positions $\mathcal{M}_t^{\mathrm{anch}}=\{l\in\bar{\mathcal{A}} 
\mid z_t^l=m\}$, i.e., $\mathcal{M}_t$ is replaced by 
$\mathcal{M}_t^{\mathrm{anch}}$ in 
Equation~\eqref{eq:app_mdlm_ct}.
The denoiser is trained to reconstruct only the missing 
coordinates conditional on the observed ones, which is exactly 
the imputation objective $p_\theta(x_0^{\mathcal{M}}\mid 
x_0^{\mathcal{O}})$ targeted in 
Appendix~\ref{app:dm_cond}.

\section{Derivation of the Soft-Label Loss Decomposition}
\label{app:soft_label}

We provide the derivation of the decomposition stated in Eq.~\ref{eq:decomposition}. For brevity, we suppress the position index $l$ and the timestep weight $w(t)$, and consider a single per-token loss
\begin{equation}
    \mathcal{L} = -\big\langle \mathbf{s}, \log \mathbf{p}_\theta \big\rangle = -\sum_{i=1}^{K} s_i \log p_\theta(i),
\end{equation}
where $\mathbf{s}$ is the ordinal-aware soft label centered at the ground-truth index $y$ (Eq.~\ref{eq:soft_label}) and $\mathbf{p}_\theta \in \Delta^{K-1}$ is the predicted distribution.

\paragraph{Step 1: Decomposition of the soft label.}
Let $\epsilon := 1 - s_y \in [0, 1)$ denote the total mass that $\mathbf{s}$ assigns outside the ground-truth index $y$. We define
\begin{equation}
    \mathbf{u} := \frac{\mathbf{s} - s_y \mathbf{e}_y}{1 - s_y},
    \label{eq:u_def}
\end{equation}
where $\mathbf{e}_y$ denotes the one-hot vector at position $y$. By construction, $u_y = 0$ and $\mathbf{u}$ is a valid probability distribution supported on the ordinal neighborhood $\{i : 0 < |i - y| \le w\}$. The soft label can then be written as a convex combination
\begin{equation}
    \mathbf{s} = (1 - \epsilon)\,\mathbf{e}_y + \epsilon\, \mathbf{u}.
    \label{eq:s_decomp}
\end{equation}

\paragraph{Step 2: Splitting the cross-entropy.}
Substituting Eq.~\ref{eq:s_decomp} into $\mathcal{L}$ and applying the linearity of the inner product gives
\begin{align}
    \mathcal{L} 
    &= -\big\langle (1-\epsilon)\,\mathbf{e}_y + \epsilon\, \mathbf{u},\ \log \mathbf{p}_\theta \big\rangle \\
    &= -(1-\epsilon)\,\langle \mathbf{e}_y, \log \mathbf{p}_\theta \rangle - \epsilon\, \langle \mathbf{u}, \log \mathbf{p}_\theta \rangle.
    \label{eq:split}
\end{align}
The first term is the standard MDLM cross-entropy against the one-hot target,
\begin{equation}
    -\langle \mathbf{e}_y, \log \mathbf{p}_\theta \rangle = -\log p_\theta(y) = \mathcal{L}_{\mathrm{NELBO}},
\end{equation}
recovering the per-token NELBO contribution of vanilla MDLM~\cite{sahoo2024simple}.

\paragraph{Step 3: KL form of the second term.}
The second term is the cross-entropy $H(\mathbf{u}, \mathbf{p}_\theta)$, which by the standard identity decomposes as
\begin{equation}
    H(\mathbf{u}, \mathbf{p}_\theta) = -\langle \mathbf{u}, \log \mathbf{p}_\theta \rangle = D_{\mathrm{KL}}(\mathbf{u}\,\|\,\mathbf{p}_\theta) + H(\mathbf{u}),
\end{equation}
where $H(\mathbf{u}) = -\sum_i u_i \log u_i$ is the entropy of $\mathbf{u}$ and depends only on $\mathbf{s}$ (i.e., independent of $\mathbf{p}_\theta$).

\paragraph{Step 4: Combining.}
Substituting back into Eq.~\ref{eq:split} yields
\begin{equation}
    \mathcal{L} = (1 - \epsilon)\,\mathcal{L}_{\mathrm{NELBO}} + \epsilon\, D_{\mathrm{KL}}(\mathbf{u}\,\|\,\mathbf{p}_\theta) + \underbrace{\epsilon\, H(\mathbf{u})}_{\mathrm{const}},
\end{equation}
which is exactly the decomposition in Eq.~\ref{eq:decomposition}. Two limiting cases are worth noting:
\begin{itemize}
    \item As the soft label sharpens ($\mathbf{s} \to \mathbf{e}_y$, equivalently $\epsilon \to 0$), the KL term vanishes and $\mathcal{L} \to \mathcal{L}_{\mathrm{NELBO}}$, recovering the vanilla MDLM objective.
    \item For $\epsilon > 0$, the additional non-negative term $\epsilon\, D_{\mathrm{KL}}(\mathbf{u}\,\|\,\mathbf{p}_\theta)$ penalizes probability mass placed outside the ordinal neighborhood of $y$, since $\mathbf{u}$ is supported only on $\{i : 0 < |i - y| \le w\}$.
\end{itemize}
The same decomposition holds inside the expectation $\mathbb{E}_t[w(t) \cdot]$ in Eq.~\ref{eq:diff_loss} by linearity, giving the full-objective form
\begin{equation}
    \mathcal{L}_{\text{diff}} 
    = \mathbb{E}_t\!\left[w(t) \sum_{l \in \mathcal{M}_t} \Big( (1-\epsilon^{(l)})\,\mathcal{L}_{\mathrm{NELBO}}^{(l)} + \epsilon^{(l)}\, D_{\mathrm{KL}}\!\big(\mathbf{u}^{(l)}\,\|\,\mathbf{p}_\theta(\mathbf{z}_t^{(l)})\big)\Big)\right] + \mathrm{const}.
\end{equation}

\section{Experimental Settings}

\subsection{Datasets}

\begin{table}[ht!]
\centering
\renewcommand{\arraystretch}{1.15}
\caption{Statistics of benchmark datasets.} \label{table:dataset_summary}
\begin{tabular}{c|cccc}
\hline
\hline
\rowcolor{gray!20}
\textbf{Datasets} & \textbf{Features} & \textbf{Frequency} & \textbf{Samples} & \textbf{Domain} \\ \hline
Energy & 28 & 10 min. & 19,735 & Weather \\ 
ETTh & 7 & 60 min. & 17,420 &  Temperature \\  
Weather & 21 & 10 min. & 52,696 & Weather \\ 
Sine & 5 & - & 10000 & Simulation \\
\hline \hline
\end{tabular}
\end{table}

We evaluate the proposed MDTIM on four time-series datasets: three real-world benchmarks (Energy, ETTh, Weather) to assess practical validity, and one synthetic Sine dataset for controlled evaluation of periodic dynamics. These datasets cover a broad range of dynamics observed in long-horizon multivariate time-series. The Energy\footnote{\url{https://archive.ics.uci.edu/dataset/374/appliances+energy+prediction}} dataset contains electricity-related measurements with strong seasonality and occasional irregular variations. The ETTh\footnote{\url{https://github.com/zhouhaoyi/ETDataset}} dataset is a standard benchmark derived from Electricity Transformer Dataset \citep{zhou2021informer} and is commonly used to evaluate long-range dependency modeling. The Weather\footnote{\url{https://www.bgc-jena.mpg.de/wetter/}} \citep{wu2021autoformer} dataset includes multivariate meteorological observations characterized by nonlinear interactions, seasonal trends, and stochastic fluctuations. In addition, the synthetic Sine dataset offers a controlled environment with known periodic structure, which helps validate fundamental forecasting behavior under minimal uncontrolled factors. Overall, these datasets were selected to ensure domain diversity and to evaluate performance under both structured periodic signals and noisy, non-stationary real-world patterns.
Dataset specifications are summarized in Table~\ref{table:dataset_summary}.
For the implementation and training of the baseline models, we adopted the codebase curated by \citet{du2024tsi}\footnote{\url{https://github.com/WenjieDu/Awesome_Imputation}}.

\subsection{Training Configurations}

We provide the detailed hyperparameter settings used to train MDTIM. To ensure reproducibility, we list the common architectural parameters and training schemes applied across all datasets. All experiments are conducted with dual Intel Xeon Gold 6444Y CPUs and a single NVIDIA H100 PCIe GPU (80GB).

\subsubsection{Hyperparameters}
\label{app:hyper}

The architecture of MDTIM is based on the Factorized Temporal-Variate Backbone described in Section \ref{sec:framework}. Table~\ref{tab:hyperparameters} summarizes detailed hyperparameters. Every architectural and discretization hyperparameter, including $K$, $w$, and $\sigma$, is shared across all datasets and missing rates. $K$ was selected once from the ETTh sweep in Table~\ref{tab:ablation_bins}, and $w$ and $\sigma$ were fixed a priori without tuning, so none of the reported gains come from per-dataset configuration.

\begin{table}[h!]
    \centering
    \renewcommand{\arraystretch}{1.15}
    \caption{Hyperparameters for MDTIM architecture and training.}
    \label{tab:hyperparameters}
    \begin{small}
    \begin{tabular}{l|l|c|l}
        \hline
        \hline
        \rowcolor{gray!20}
        \textbf{Category} & \textbf{Parameter} & \textbf{Value} & \textbf{Description} \\
        \hline
        \multirow{7}{*}{\textbf{Model}} 
          & Hidden Size ($D$) & 256 & Dimension of hidden states \\
          & Attention Heads & 16 & Number of heads in MSA \\
          & DiT Blocks ($L$) & 5 & Number of factorized layers \\
          & Dropout & 0.2 & Dropout probability \\
          & Bins ($K$) & 40 & Vocabulary size for discretization \\
          & Soft-label window ($w$) & 2 & Truncation window of Eq.~\ref{eq:soft_label} \\
          & Soft-label smoothness ($\sigma$) & 1.0 & Gaussian width in token bins \\
          & Inference passes ($M$) & 10 & Noise samples averaged in Eq.~\ref{eq:expectation} \\
          & Output Range & 1.5 & Value range $[-1.5, 1.5]$ \\
          & Cond. Dim & 16 & Time-step embedding dimension \\
        \hline
        \multirow{8}{*}{\textbf{Training}} 
          & Batch Size & 256 & - \\
          & Learning Rate & $3 \times 10^{-4}$ & - \\
          & Scheduler & Const. w/ Warmup & - \\
          & Warmup Steps & 2,500 & - \\
          & Max Steps & 10,000 & - \\
          & Gradient Clip & 1.0 & Norm clipping value \\
          & EMA Decay & 0.995 & Exponential moving average \\
          & FFT Weight ($\lambda$) & 1.0 & Spectral loss weight \\
        \hline \hline
    \end{tabular}
    \end{small}
\end{table}

\subsubsection{Noise Schedule and Time-Dependent Weighting}
\label{noise}
We employ a Log-Linear Noise Schedule to compute the continuous-time diffusion process. For a time step $t \in [0, 1]$ and a small constant $\epsilon = 10^{-3}$, the total noise $\sigma(t)$ and its rate of change are defined as:

\begin{equation}
    \sigma(t) = -\log\left(1 - (1-\epsilon)t\right), \quad \frac{d\sigma}{dt} = \frac{1-\epsilon}{1-(1-\epsilon)t}
\end{equation}

To ensure balanced training across varying noise levels, we apply importance sampling via a time-dependent loss weight $w(t)$. This weight effectively normalizes the contribution of each time step to the objective function:

\begin{equation}
    w(t) = \frac{d\sigma(t)/dt}{e^{\sigma(t)} - 1}
\end{equation}

This weighting scheme assigns higher importance to low-noise regions ($t \approx 0$), prioritizing the learning of fine-grained details from clean data, while down-weighting highly corrupted states ($t \approx 1$) where reconstruction is ambiguous.

\subsection{Evaluation Metrics}
\label{appendix:metrics}

We evaluate both point imputation accuracy and the quality of predictive distributions.
Let $y_{i,t,c}$ and $\hat{y}_{i,t,c}$ denote the ground truth and the imputed value for sample $i$, time step $t$, and channel $c$.
All metrics are computed exclusively on the masked positions, denoted by the set of indices $\mathcal{M}$.

\paragraph{MAE and MSE.}
We report Mean Absolute Error (MAE) and Mean Squared Error (MSE), averaged over the masked entries $\mathcal{M}$:
\begin{equation}
\mathrm{MAE}
=
\frac{1}{|\mathcal{M}|}\sum_{(i,t,c)\in\mathcal{M}}
\left| \hat{y}_{i,t,c}-y_{i,t,c}\right|,
\qquad
\mathrm{MSE}
=
\frac{1}{|\mathcal{M}|}\sum_{(i,t,c)\in\mathcal{M}}
\left( \hat{y}_{i,t,c}-y_{i,t,c}\right)^2.
\end{equation}
In probabilistic settings, we set $\hat{y}_{i,t,c}$ to the sample mean of $M$ generated trajectories,
$\hat{y}_{i,t,c}=\frac{1}{M}\sum_{m=1}^{M}\tilde{y}_{i,t,c}^{(m)}$.

\paragraph{CRPS.}
To assess the probabilistic imputation performance, we use the Continuous Ranked Probability Score (CRPS)~\citep{matheson1976scoring}.
For a predictive CDF $F_{i,t,c}$ at a masked position, CRPS is defined as
\begin{equation}
\mathrm{CRPS}\!\left(F_{i,t,c},y_{i,t,c}\right)
=
\int_{-\infty}^{\infty}
\Big(F_{i,t,c}(z)-\mathbb{I}[z\ge y_{i,t,c}]\Big)^2\,dz.
\end{equation}
Lower CRPS indicates better probabilistic imputation, rewarding both calibration and sharpness.

\paragraph{Sample-based CRPS Estimation.}
Since the model produces $M$ stochastic samples $\{\tilde{y}_{i,t,c}^{(m)}\}_{m=1}^{M}\sim F_{i,t,c}$, we approximate CRPS by
\begin{equation}
\overline{\mathrm{CRPS}}_{i,t,c}
=
\frac{1}{M}\sum_{m=1}^{M}\left|\tilde{y}_{i,t,c}^{(m)}-y_{i,t,c}\right|
-
\frac{1}{2M^{2}}\sum_{m=1}^{M}\sum_{n=1}^{M}\left|\tilde{y}_{i,t,c}^{(m)}-\tilde{y}_{i,t,c}^{(n)}\right|.
\end{equation}
We report the average of $\overline{\mathrm{CRPS}}_{i,t,c}$ over all masked entries in $\mathcal{M}$.

\section{Additional Experimental Results}
\label{app:additional_results}

In this section, we provide supplementary experimental results that complement the main paper. Section~\ref{app:full_results} reports the full quantitative comparison including the 50\% missing ratio omitted from Table~\ref{tab:main_results1} for space. Section~\ref{app:continuous_ablation} presents the two controls behind the missingness-mechanism ablation of Section~\ref{sec:ablation}, Section~\ref{app:shared} examines a channel-shared embedding, and Section~\ref{app:seeds} reports variability across training seeds.

\subsection{Full Imputation Performance Across All Missing Ratios}
\label{app:full_results}

Table~\ref{tab:full_results} reports the complete results of our main imputation experiment, including the 30\%, 50\%, and 70\% missing ratios under both Uniform and Geometric scenarios. The 50\% column extends the trend reported in the main paper: MDTIM consistently maintains the lowest MAE across the majority of settings, and the relative improvement over baselines remains stable across missing ratios. This confirms that the performance advantage of MDTIM is not specific to any particular corruption level, but holds robustly throughout the regime of partial observation.

\begin{table}[h!]
\setlength{\tabcolsep}{3pt}
\centering
\renewcommand{\arraystretch}{1.13}
\caption{Quantitative comparison of multivariate time-series imputation performance ($L=48$). We report the MAE of MDTIM and baselines averaged over 3 random mask seeds. The best results are highlighted in \textbf{bold}.}
\label{tab:full_results}
\resizebox{1\columnwidth}{!}{%
\begin{tabular}{l|l|l|l|l|l|l|l|l|l|l|l|l|l|l|l|l|l|l|l|l|l|l|l|l|l}
\hline
\hline
\rowcolor{gray!20}
\multicolumn{2}{c|}{\textbf{Dataset}}
&\multicolumn{6}{c|}{\textbf{Energy}}&\multicolumn{6}{c|}{\textbf{ETTh}}&\multicolumn{6}{c|}{\textbf{Weather}}&\multicolumn{6}{c}{\textbf{Sine}}\\ 
\hline
\rowcolor{gray!20}
\multicolumn{2}{c|}{\textbf{Missing Type}}
&\multicolumn{3}{c|}{\textbf{Uniform}}&\multicolumn{3}{c|}{\textbf{Geometric}}&\multicolumn{3}{c|}{\textbf{Uniform}}&\multicolumn{3}{c|}{\textbf{Geometric}}&\multicolumn{3}{c|}{\textbf{Uniform}}&\multicolumn{3}{c|}{\textbf{Geometric}}&\multicolumn{3}{c|}{\textbf{Uniform}}&\multicolumn{3}{c}{\textbf{Geometric}}\\ 
\hline
\rowcolor{gray!20}
\multicolumn{2}{c|}{Model}
&30\%&50\%&70\%&30\%&50\%&70\%&30\%&50\%&70\%&30\%&50\%&70\%&30\%&50\%&70\%&30\%&50\%&70\%&30\%&50\%&70\%&30\%&50\%&70\%\\ 
\hline
\multirow{3}{*}{\begin{turn}{90}RNN\end{turn}}
  & BRITS                    & 0.246          & 0.296          & 0.379          & 0.329          & 0.344          & 0.366          & 0.194          & 0.235          & 0.299          & 0.227          & 0.258          & 0.292          & 0.050          & 0.057          & 0.073          & 0.058          & 0.063          & 0.071          & 0.010          & 0.013          & 0.021          & 0.020          & 0.019          & 0.019          \\
 & MRNN                     & 1.086          & 1.107          & 1.143          & 1.087          & 1.108          & 1.143          & 0.743          & 0.762          & 0.782          & 0.751          & 0.765          & 0.781          & 0.651          & 0.653          & 0.661          & 0.654          & 0.656          & 0.659          & 0.169          & 0.169          & 0.170          & 0.169          & 0.169          & 0.170          \\
 & GRUD                     & 0.364          & 0.387          & 0.457          & 0.426          & 0.425          & 0.445          & 0.310          & 0.336          & 0.394          & 0.348          & 0.362          & 0.383          & 0.104          & 0.176          & 0.369          & 0.164          & 0.231          & 0.350          & 0.008          & 0.008          & 0.014          & 0.013          & 0.012          & 0.012          \\ \hline
\multirow{5}{*}{\begin{turn}{90}Transformer\end{turn}}
 & Transformer              & 0.323          & 0.335          & 0.401          & 0.354          & 0.356          & 0.393          & 0.168          & 0.201          & 0.252          & 0.183          & 0.212          & 0.247          & 0.076          & 0.067          & 0.070          & 0.083          & 0.072          & 0.068          & 0.039          & 0.041          & 0.046          & 0.042          & 0.044          & 0.045          \\
 & Informer                 & 0.344          & 0.338          & 0.379          & 0.371          & 0.357          & 0.373          & 0.207          & 0.240          & 0.302          & 0.226          & 0.256          & 0.296          & 0.047          & 0.046          & 0.059          & 0.052          & 0.050          & 0.058          & 0.027          & 0.025          & 0.039          & 0.036          & 0.032          & 0.037          \\
 & PatchTST                 & 0.586          & 0.336          & 0.529          & 0.528          & 0.378          & 0.517          & 0.202          & 0.218          & 0.272          & 0.231          & 0.243          & 0.262          & 0.074          & 0.061          & 0.082          & 0.081          & 0.069          & 0.077          & 0.015          & 0.011          & 0.017          & 0.019          & 0.014          & 0.015          \\
 & SAITS                    & 0.177          & 0.185          & 0.212          & 0.204          & 0.200          & 0.208          & \underline{0.140} & \underline{0.166} & \textbf{0.210}    & \underline{0.152}    & \underline{0.176}    & \underline{0.206} & 0.045          & 0.044          & 0.049          & 0.050          & 0.048          & 0.048          & 0.026          & 0.023          & 0.026          & 0.030          & 0.026          & 0.025          \\
 & Imputeformer & 0.066 & 0.100           & 0.219   & 0.113 & 0.143 & 0.194     & 0.146 & 0.178 & 0.245 & 0.165 & 0.196 & 0.235 & 0.049 & 0.068 & 0.171 & 0.090  & 0.111 & 0.149 & 0.006 & 0.012 & 0.041 & 0.012 & 0.019 & 0.037 \\
 \hline
\multirow{2}{*}{\begin{turn}{90}CNN\end{turn}}
 & TimesNet                 & 0.617          & 0.683          & 0.818          & 0.689          & 0.726          & 0.808          & 0.593          & 0.653          & 0.719          & 0.628          & 0.666          & 0.717          & 0.211          & 0.202          & 0.351          & 0.225          & 0.223          & 0.348          & 0.167          & 0.188          & 0.220          & 0.189          & 0.200          & 0.217          \\
 & SCINet                   & 0.557          & 0.418          & 0.571          & 0.546          & 0.442          & 0.565          & 0.238          & 0.252          & 0.323          & 0.261          & 0.273          & 0.317          & 0.074          & 0.068          & 0.089          & 0.086          & 0.076          & 0.087          & 0.019          & 0.018          & 0.031          & 0.026          & 0.024          & 0.029          \\
 \hline
\multirow{3}{*}{\begin{turn}{90}Linear\end{turn}}
 & DLinear                  & 0.795          & 0.351          & 0.527          & 0.684          & 0.388          & 0.517          & 0.379          & 0.319          & 0.445          & 0.373          & 0.357          & 0.435          & 0.372          & 0.149          & 0.223          & 0.345          & 0.161          & 0.220          & 0.070          & 0.032          & 0.053          & 0.062          & 0.041          & 0.051          \\
 & FiLM                     & 0.877          & 0.380          & 0.520          & 0.791          & 0.413          & 0.512          & 0.696          & 0.608          & 0.627          & 0.707          & 0.627          & 0.623          & 0.380          & 0.156          & 0.209          & 0.360          & 0.165          & 0.206          & 0.127          & 0.102          & 0.109          & 0.132          & 0.110          & 0.108          \\
 & FreTS                    & 0.157          & 0.147          & 0.219          & 0.225          & 0.184          & 0.206          & 0.222          & 0.238          & 0.299          & 0.262          & 0.266          & 0.286          & 0.071          & 0.059          & 0.080          & 0.080          & 0.067          & 0.074          & 0.068          & 0.037          & 0.068          & 0.075          & 0.049          & 0.065          \\  \hline
\multirow{5}{*}{\begin{turn}{90} Generative \end{turn}}
 & GPVAE                    & 0.476          & 0.509          & 0.730          & 0.501          & 0.532          & 0.727          & 0.333          & 0.349          & 0.449          & 0.369          & 0.385          & 0.440          & 0.158          & 0.153          & 0.255          & 0.172          & 0.170          & 0.252          & 0.152          & 0.150          & 0.162          & 0.153          & 0.151          & 0.162          \\
 & USGAN                    & 0.264          & 0.323          & 0.412          & 0.330          & 0.362          & 0.401          & 0.208          & 0.243          & 0.302          & 0.240          & 0.266          & 0.295          & 0.087          & 0.094          & 0.122          & 0.101          & 0.104          & 0.118          & 0.014          & 0.018          & 0.029          & 0.026          & 0.026          & 0.026          \\ 
 & CSDI                     & 0.094    & 0.112    & 0.139    & 0.110    & 0.122    & 0.136    & 0.160          & 0.192          & 0.244          & 0.177          & 0.206          & 0.240          & 0.039    & \underline{0.043}    & \underline{0.049}    & \underline{0.044}    & \underline{0.046}    & \underline{0.048}    & 0.003 & 0.004  & 0.004    & 0.004    & 0.004    & \underline{0.004} \\
 & FGTI                     & \underline{0.050}  & \underline{0.071}         & \underline{0.100}     & \underline{0.065} & \underline{0.080}  & \underline{0.097}     & 0.218 & 0.255 & 0.349 & 0.293 & 0.311 & 0.328 & \underline{0.038} & \underline{0.043} & 0.051 & 0.046 & 0.047 & 0.049 & \textbf{0.001}     & \textbf{0.002} & \textbf{0.003} & \textbf{0.002} & \textbf{0.003} & \textbf{0.003} \\
 & MDTIM (Ours)    & \textbf{0.044} & \textbf{0.061} & \textbf{0.085}          & \textbf{0.053} & \textbf{0.067} & \textbf{0.082}  & \textbf{0.127} & \textbf{0.157} & \underline{0.211} & \textbf{0.146}    & \textbf{0.173}    & \textbf{0.205}  & \textbf{0.032} & \textbf{0.036} & \textbf{0.044}   & \textbf{0.036} & \textbf{0.039} & \textbf{0.043}  & \underline{0.002} & \underline{0.003} & \underline{0.004} & \underline{0.003} & \textbf{0.003} & \textbf{0.003} 
 \\ \hline  \hline
\end{tabular} 
}
\end{table}

\subsection{Comparison with Continuous Modeling}
\label{app:continuous_ablation}

To isolate the contribution of our discrete formulation, we compare MDTIM (Disc), the proposed model, against a continuous-modeling variant MDTIM (Cont). MDTIM (Cont) shares the same Factorized Temporal-Variate backbone and masked diffusion training paradigm, but operates directly on continuous input values and is optimized with a Mean Squared Error (MSE) loss. In contrast, MDTIM (Disc) employs our \textit{Stochastic Discretization} to map continuous signals into discrete tokens and optimizes the ordinal-aware discrete diffusion objective described in Section~\ref{sec:framework}.

\begin{figure}[h!]
    \centering
    \begin{subfigure}[b]{0.35\columnwidth}
        \centering
        \includegraphics[width=\linewidth]{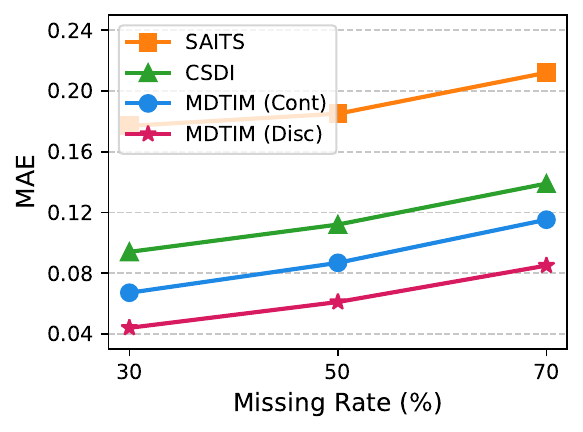}
        \caption{Uniform Missing}
        \label{fig:ablation_uni}
    \end{subfigure}
    \hspace{0.05\columnwidth}
    \begin{subfigure}[b]{0.35\columnwidth}
        \centering
        \includegraphics[width=\linewidth]{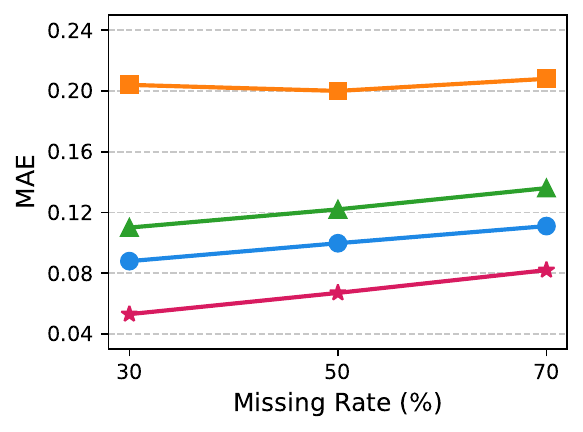}
        \caption{Geometric Missing}
        \label{fig:ablation_geo}
    \end{subfigure}
    \caption{Comparison between MDTIM (Disc) and MDTIM (Cont) on the Energy dataset, alongside representative baselines (SAITS, CSDI). Results are reported under both uniform and geometric missing scenarios across varying missing rates.}
    \label{fig:ablation}
\end{figure}

As shown in Figure~\ref{fig:ablation}, MDTIM (Cont) already performs competitively with strong baselines, indicating that the masked diffusion paradigm of learning to reconstruct from varying corruption levels with timestep conditioning is itself effective even in continuous spaces. Nevertheless, MDTIM (Disc) consistently achieves lower MAE across all missing rates and missing types. This performance gap confirms that the explicit structural separation between valid observations and missing placeholders, achieved through our orthogonal tokenization, provides a benefit beyond the masked diffusion training scheme alone, and is crucial for time series imputation.

\paragraph{Attention-mask variant.} The second control in Section~\ref{sec:ablation} replaces the \texttt{[MASK]} token by an attention mask. The input still carries an embedding at every position, but positions that are missing are removed from the key set of both the temporal and the variate attention in every block, so each query attends only to observed positions and to itself. Missing positions therefore read the observed ones, while no observed position can see where the gaps are. Everything else, including the discretization, the soft-label objective, the spectral term, and the training schedule, is identical to MDTIM, and the variant is evaluated under the same three mask seeds.

\subsection{Channel-Shared Embedding}
\label{app:shared}
The per-channel embedding tables $E_c$ of Section~\ref{sec:framework} hold $CKD$ parameters, a cost that grows linearly with the number of variates. To see how much of the accuracy depends on them, we retrain MDTIM on Energy with a single embedding table shared across all channels, which costs $KD$ parameters, a 28-fold reduction, with everything else identical.

\begin{table}[ht]
\centering
\caption{MAE on Energy with per-channel and channel-shared embedding tables, averaged over 3 random mask seeds.}
\label{tab:shared_embed}
\begin{tabular}{l|ccc|ccc}
\hline
\hline
\rowcolor{gray!20}
& \multicolumn{3}{c|}{\textbf{Uniform}} & \multicolumn{3}{c}{\textbf{Geometric}} \\
\hline
\rowcolor{gray!20}
Embedding & 30\% & 50\% & 70\% & 30\% & 50\% & 70\% \\
\hline
Per-channel ($CKD$) & 0.044 & 0.061 & 0.085 & 0.053 & 0.067 & 0.082 \\
Shared ($KD$)       & 0.044 & 0.061 & 0.085 & 0.053 & 0.067 & 0.082 \\
\hline
\hline
\end{tabular}
\end{table}

The two variants perform identically to three decimals in every setting (Table~\ref{tab:shared_embed}). The RoPE attention that captures temporal correlation also picks up the cross-channel structure, which makes the shared table a practical option for data with many channels. We would be careful about extrapolating this at much higher $C$, since Energy has only 28 channels, still far fewer than the several hundred of a truly high-dimensional dataset.

\subsection{Variance Across Training Seeds}
\label{app:seeds}
The mask seed determines which entries are missing, so changing it changes the imputation problem itself, and a standard deviation taken across mask seeds would blend problem difficulty with model variability. We therefore report variability across training seeds under a fixed mask. For each of the three mask seeds used in Section~\ref{sec:experiments}, we train three models with seeds 2024, 2025, and 2026 and report the mean and standard deviation of their MAE (Table~\ref{tab:seeds}). The training-seed standard deviation stays at or below 0.001 in every setting, so the comparisons in the main text are not an artifact of seed noise.

\begin{table}[ht]
\centering
\caption{MAE mean $\pm$ standard deviation over three training seeds (2024, 2025, 2026), reported separately for each mask seed.}
\label{tab:seeds}
\resizebox{\columnwidth}{!}{%
\begin{tabular}{cc|ccc|ccc}
\hline
\hline
\rowcolor{gray!20}
& & \multicolumn{3}{c|}{\textbf{Uniform}} & \multicolumn{3}{c}{\textbf{Geometric}} \\
\hline
\rowcolor{gray!20}
Dataset & Rate & Mask 2024 & Mask 2025 & Mask 2026 & Mask 2024 & Mask 2025 & Mask 2026 \\
\hline
\multirow{3}{*}{ETTh} & 30\% & 0.1270$\pm$0.0005 & 0.1274$\pm$0.0006 & 0.1269$\pm$0.0007 & 0.1461$\pm$0.0004 & 0.1449$\pm$0.0005 & 0.1456$\pm$0.0005 \\
 & 50\% & 0.1569$\pm$0.0001 & 0.1578$\pm$0.0002 & 0.1569$\pm$0.0003 & 0.1716$\pm$0.0001 & 0.1733$\pm$0.0003 & 0.1721$\pm$0.0003 \\
 & 70\% & 0.2118$\pm$0.0009 & 0.2113$\pm$0.0009 & 0.2110$\pm$0.0010 & 0.2045$\pm$0.0009 & 0.2048$\pm$0.0008 & 0.2056$\pm$0.0006 \\
\hline
\multirow{3}{*}{Energy} & 30\% & 0.0439$\pm$0.0001 & 0.0445$\pm$0.0001 & 0.0441$\pm$0.0001 & 0.0527$\pm$0.0002 & 0.0546$\pm$0.0001 & 0.0532$\pm$0.0002 \\
 & 50\% & 0.0606$\pm$0.0001 & 0.0610$\pm$0.0002 & 0.0608$\pm$0.0001 & 0.0669$\pm$0.0001 & 0.0672$\pm$0.0001 & 0.0668$\pm$0.0001 \\
 & 70\% & 0.0845$\pm$0.0002 & 0.0850$\pm$0.0002 & 0.0845$\pm$0.0001 & 0.0819$\pm$0.0001 & 0.0823$\pm$0.0002 & 0.0820$\pm$0.0002 \\
\hline
\multirow{3}{*}{Weather} & 30\% & 0.0316$\pm$0.0001 & 0.0316$\pm$0.0001 & 0.0317$\pm$0.0001 & 0.0360$\pm$0.0001 & 0.0363$\pm$0.0000 & 0.0363$\pm$0.0000 \\
 & 50\% & 0.0360$\pm$0.0000 & 0.0360$\pm$0.0001 & 0.0361$\pm$0.0000 & 0.0387$\pm$0.0001 & 0.0390$\pm$0.0000 & 0.0390$\pm$0.0000 \\
 & 70\% & 0.0440$\pm$0.0001 & 0.0441$\pm$0.0001 & 0.0440$\pm$0.0001 & 0.0427$\pm$0.0001 & 0.0429$\pm$0.0001 & 0.0428$\pm$0.0001 \\
\hline
\multirow{3}{*}{Sine} & 30\% & 0.0021$\pm$0.0002 & 0.0021$\pm$0.0002 & 0.0021$\pm$0.0002 & 0.0026$\pm$0.0003 & 0.0027$\pm$0.0003 & 0.0027$\pm$0.0003 \\
 & 50\% & 0.0024$\pm$0.0002 & 0.0025$\pm$0.0002 & 0.0024$\pm$0.0002 & 0.0027$\pm$0.0002 & 0.0028$\pm$0.0002 & 0.0028$\pm$0.0002 \\
 & 70\% & 0.0033$\pm$0.0002 & 0.0032$\pm$0.0002 & 0.0032$\pm$0.0002 & 0.0031$\pm$0.0002 & 0.0032$\pm$0.0002 & 0.0031$\pm$0.0002 \\
\hline
\hline
\end{tabular}
}
\end{table}

\section{Visualization of Imputation Results}
\label{appendix:qualitative}

\begin{figure}[h!]
    \centering
    \begin{subfigure}{0.49\textwidth}
        \centering
        \includegraphics[width=\linewidth]{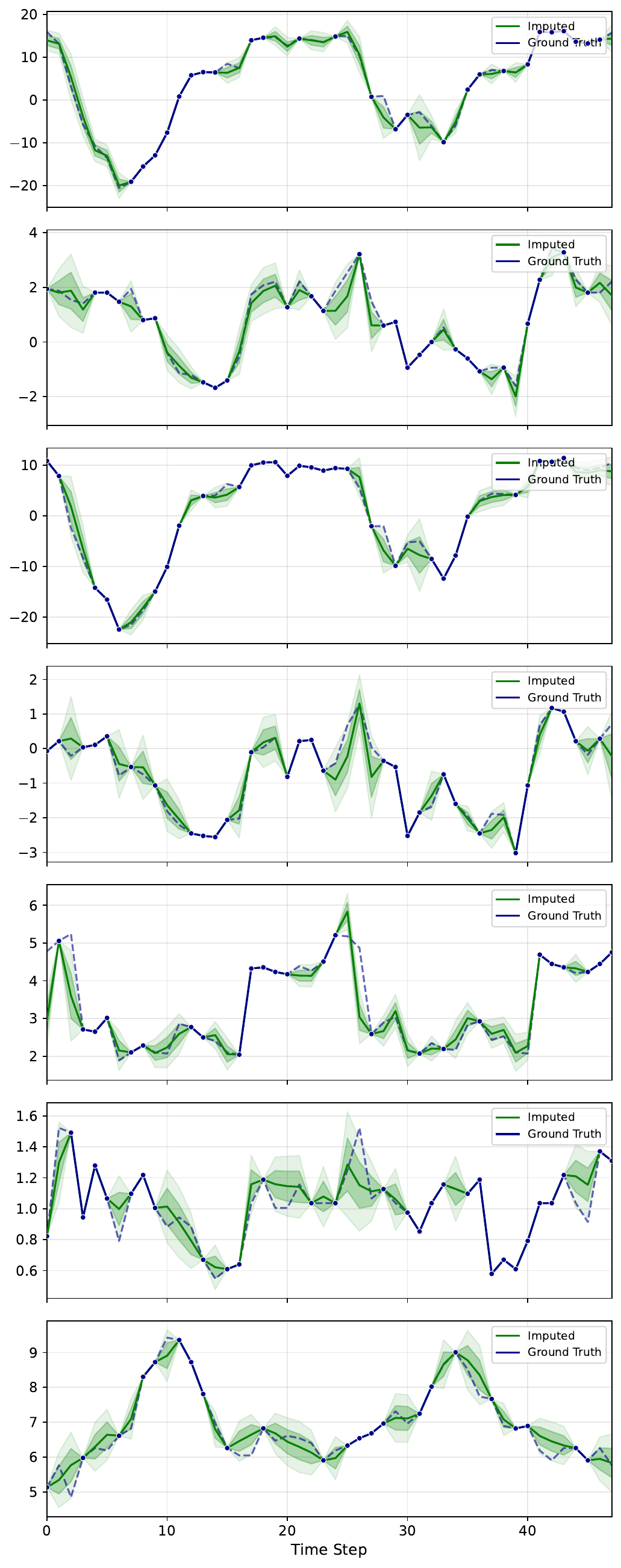}
        \caption{Uniform Missing}
    \end{subfigure}
    \hfill
    \begin{subfigure}{0.49\textwidth}
        \centering
        \includegraphics[width=\linewidth]{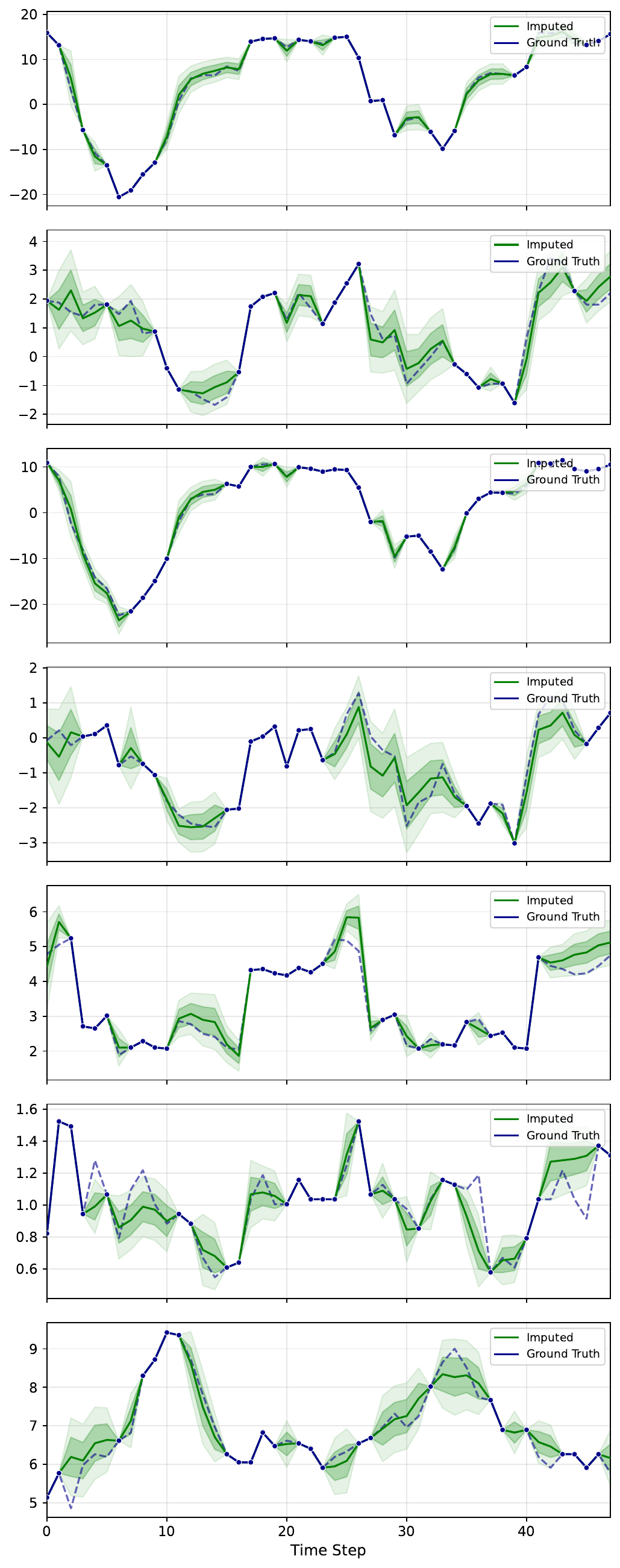}
        \caption{Geometric Missing}
    \end{subfigure}
    \caption{Visualization of imputation results on the ETTh dataset.}
    \label{fig:vis_etth}
\end{figure}

\begin{figure}[h!]
    \centering
    \begin{subfigure}{0.49\textwidth}
        \centering
        \includegraphics[width=\linewidth]{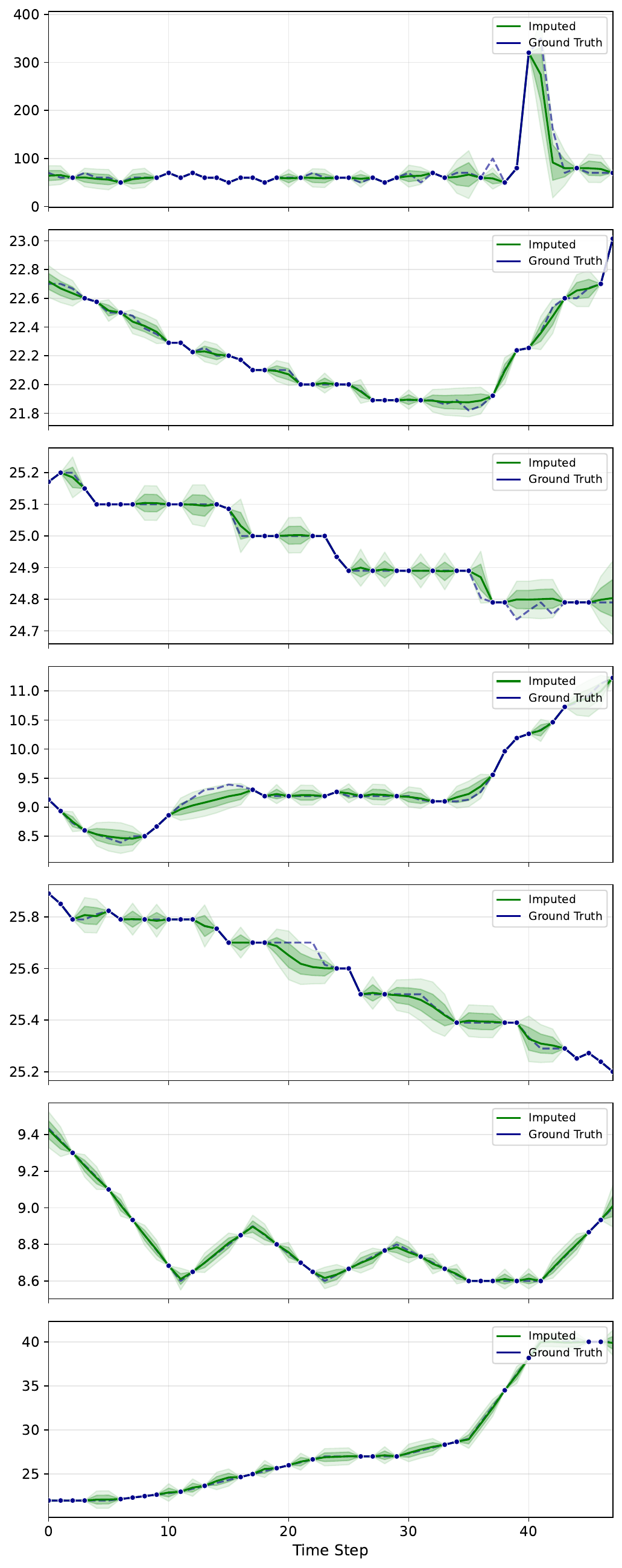}
        \caption{Uniform Missing}
    \end{subfigure}
    \hfill
    \begin{subfigure}{0.49\textwidth}
        \centering
        \includegraphics[width=\linewidth]{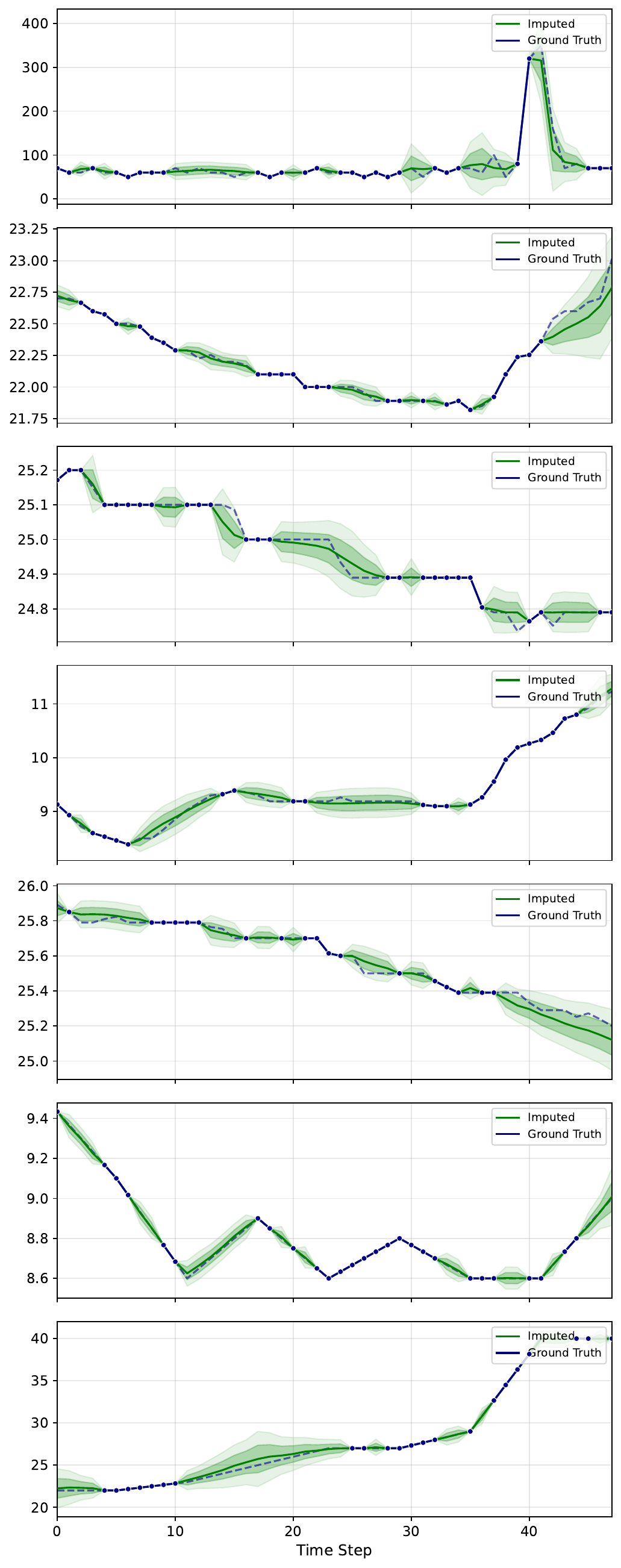}
        \caption{Geometric Missing}
    \end{subfigure}
    \caption{Visualization of imputation results on the Energy dataset.}
    \label{fig:vis_energy}
\end{figure}

\begin{figure}[h!]
    \centering
    \begin{subfigure}{0.49\textwidth}
        \centering
        \includegraphics[width=\linewidth]{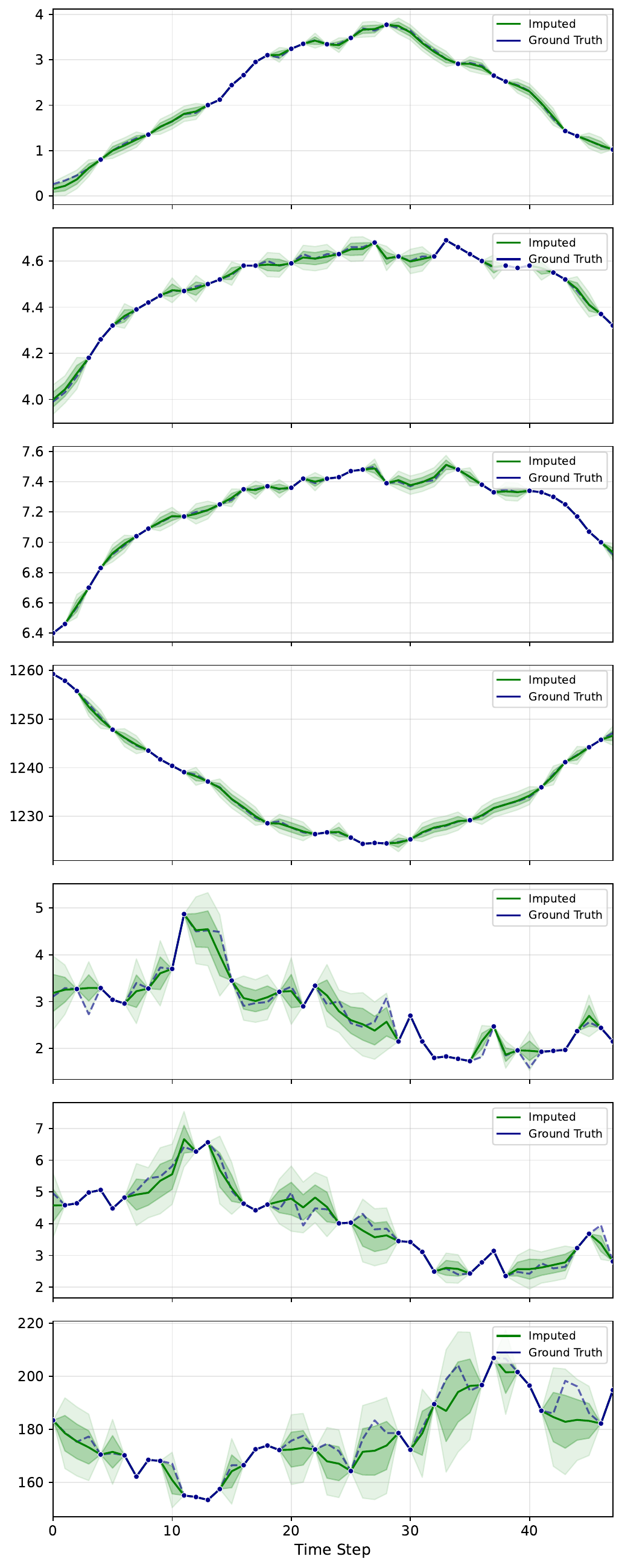}
        \caption{Uniform Missing}
    \end{subfigure}
    \hfill
    \begin{subfigure}{0.49\textwidth}
        \centering
        \includegraphics[width=\linewidth]{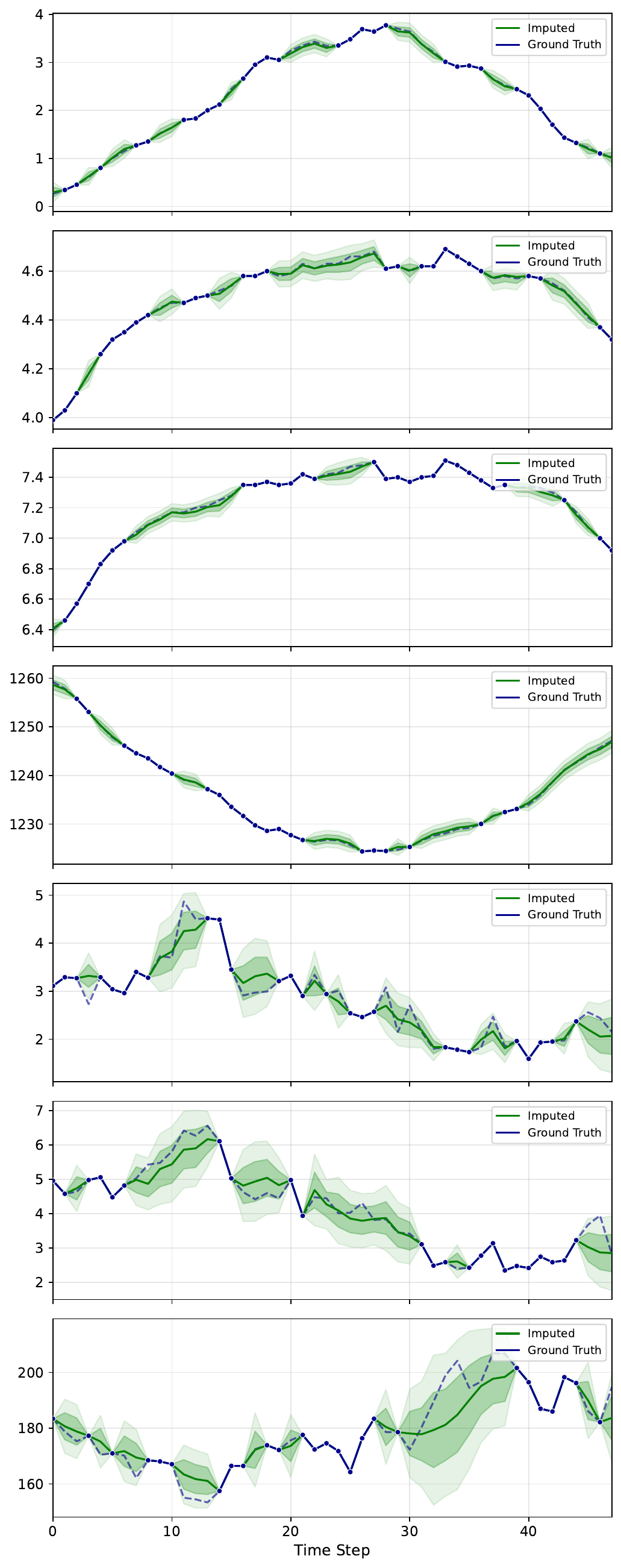}
        \caption{Geometric Missing}
    \end{subfigure}
    \caption{Visualization of imputation results on the Weather dataset.}
    \label{fig:vis_weather}
\end{figure}

\begin{figure}[h!]
    \centering
    \begin{subfigure}{0.49\textwidth}
        \centering
        \includegraphics[width=\linewidth]{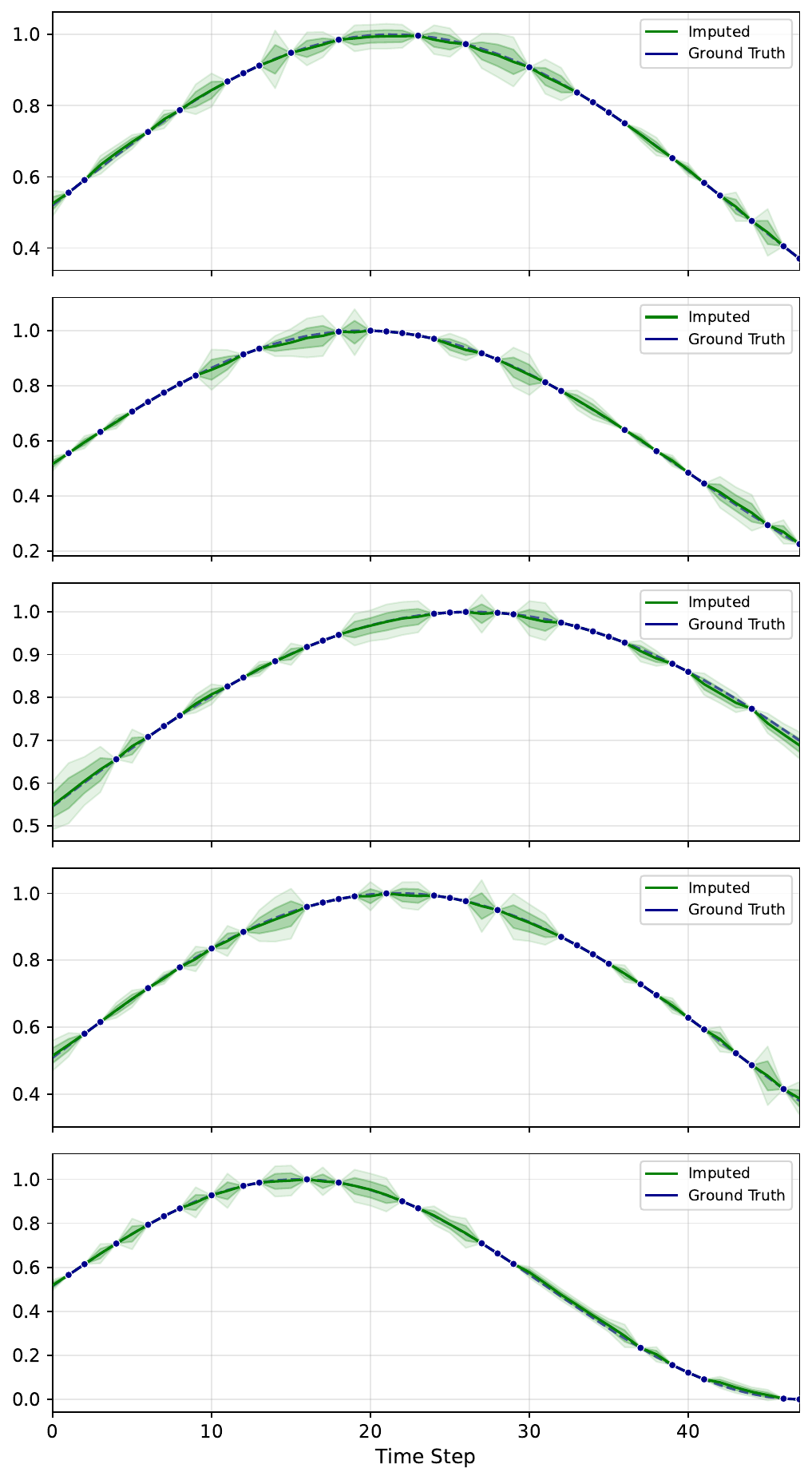}
        \caption{Uniform Missing}
    \end{subfigure}
    \hfill
    \begin{subfigure}{0.49\textwidth}
        \centering
        \includegraphics[width=\linewidth]{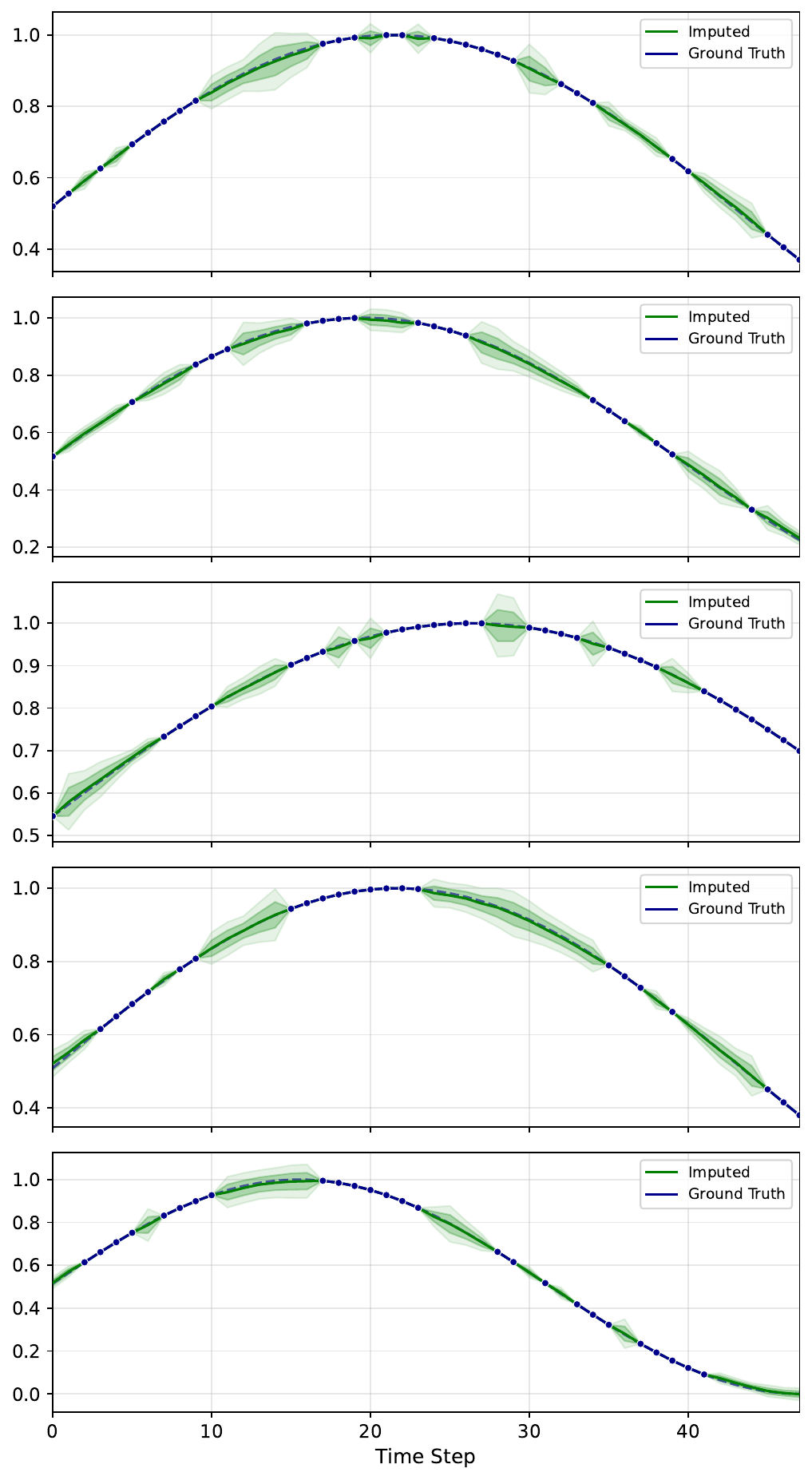}
        \caption{Geometric Missing}
    \end{subfigure}
    \caption{Visualization of imputation results on the Sine dataset.}
    \label{fig:vis_sine}
\end{figure}

We provide comprehensive visualizations of the imputation results across all benchmark datasets: ETTh, Energy, Weather, and Sine with MDTIM. 
Figures~\ref{fig:vis_etth},~\ref{fig:vis_energy},~\ref{fig:vis_weather}, and~\ref{fig:vis_sine} illustrate the reconstructed time series for all channels (ETTh, Sine) or subset of channels (Energy, Weather). 
In each figure, the left column displays the results under 50\% uniform masking, while the right column shows under 50\% geometric masking.

The blue lines represent the ground truth values, while the green lines denote the imputed values ($\hat{x}$) reconstructed by MDTIM. 
The green shaded areas indicate the estimated uncertainty intervals. 
As observed, MDTIM effectively captures the complex temporal dynamics and periodicity of the multivariate time series. 

% --- Figure: ETTh ---

% --- Figure: Energy ---

% --- Figure: Weather ---

% --- Figure: Sine ---

%%%%%%%%%%%%%%%%%%%%%%%%%%%%%%%%%%%%%%%%%%%%%%%%%%%%%%%%%%%%

\clearpage
\section*{NeurIPS Paper Checklist}

%%% END INSTRUCTIONS %%%

\begin{enumerate}

\item {\bf Claims}
    \item[] Question: Do the main claims made in the abstract and introduction accurately reflect the paper's contributions and scope?
    \item[] Answer: \answerYes{} % Replace by \answerYes{}, \answerNo{}, or \answerNA{}.
    \item[] Justification: The abstract and introduction (Section 1) clearly state our three main contributions: (i) the masked diffusion framework for time series imputation, (ii) the Stochastic Discretization and Ordinal-Aware Soft Labeling pipeline, and (iii) the Expectation-based Unmasking strategy with Spectral Consistency Regularization. Each claim is supported by experiments in Section 5.
    \item[] Guidelines:
    \begin{itemize}
        \item The answer \answerNA{} means that the abstract and introduction do not include the claims made in the paper.
        \item The abstract and/or introduction should clearly state the claims made, including the contributions made in the paper and important assumptions and limitations. A \answerNo{} or \answerNA{} answer to this question will not be perceived well by the reviewers. 
        \item The claims made should match theoretical and experimental results, and reflect how much the results can be expected to generalize to other settings. 
        \item It is fine to include aspirational goals as motivation as long as it is clear that these goals are not attained by the paper. 
    \end{itemize}

\item {\bf Limitations}
    \item[] Question: Does the paper discuss the limitations of the work performed by the authors?
    \item[] Answer: \answerYes{} % Replace by \answerYes{}, \answerNo{}, or \answerNA{}.
    \item[] Justification: We discuss limitations in Section 6: Stochastic Discretization is unbiased in expectation but lossy for an individual sample, the spectral consistency term is disabled on sparsely observed data such as PhysioNet, the per-channel embedding tables cost $O(CKD)$ parameters, and the evaluation is limited to standard imputation benchmarks of moderate length.
    \item[] Guidelines:
    \begin{itemize}
        \item The answer \answerNA{} means that the paper has no limitation while the answer \answerNo{} means that the paper has limitations, but those are not discussed in the paper. 
        \item The authors are encouraged to create a separate ``Limitations'' section in their paper.
        \item The paper should point out any strong assumptions and how robust the results are to violations of these assumptions (e.g., independence assumptions, noiseless settings, model well-specification, asymptotic approximations only holding locally). The authors should reflect on how these assumptions might be violated in practice and what the implications would be.
        \item The authors should reflect on the scope of the claims made, e.g., if the approach was only tested on a few datasets or with a few runs. In general, empirical results often depend on implicit assumptions, which should be articulated.
        \item The authors should reflect on the factors that influence the performance of the approach. For example, a facial recognition algorithm may perform poorly when image resolution is low or images are taken in low lighting. Or a speech-to-text system might not be used reliably to provide closed captions for online lectures because it fails to handle technical jargon.
        \item The authors should discuss the computational efficiency of the proposed algorithms and how they scale with dataset size.
        \item If applicable, the authors should discuss possible limitations of their approach to address problems of privacy and fairness.
        \item While the authors might fear that complete honesty about limitations might be used by reviewers as grounds for rejection, a worse outcome might be that reviewers discover limitations that aren't acknowledged in the paper. The authors should use their best judgment and recognize that individual actions in favor of transparency play an important role in developing norms that preserve the integrity of the community. Reviewers will be specifically instructed to not penalize honesty concerning limitations.
    \end{itemize}

\item {\bf Theory assumptions and proofs}
    \item[] Question: For each theoretical result, does the paper provide the full set of assumptions and a complete (and correct) proof?
    \item[] Answer: \answerNA{} % Replace by \answerYes{}, \answerNo{}, or \answerNA{}.
    \item[] Justification: The paper does not contain formal theoretical results; the contribution is methodological and empirical.
    \item[] Guidelines:
    \begin{itemize}
        \item The answer \answerNA{} means that the paper does not include theoretical results. 
        \item All the theorems, formulas, and proofs in the paper should be numbered and cross-referenced.
        \item All assumptions should be clearly stated or referenced in the statement of any theorems.
        \item The proofs can either appear in the main paper or the supplemental material, but if they appear in the supplemental material, the authors are encouraged to provide a short proof sketch to provide intuition. 
        \item Inversely, any informal proof provided in the core of the paper should be complemented by formal proofs provided in appendix or supplemental material.
        \item Theorems and Lemmas that the proof relies upon should be properly referenced. 
    \end{itemize}

    \item {\bf Experimental result reproducibility}
    \item[] Question: Does the paper fully disclose all the information needed to reproduce the main experimental results of the paper to the extent that it affects the main claims and/or conclusions of the paper (regardless of whether the code and data are provided or not)?
    \item[] Answer: \answerYes{} % Replace by \answerYes{}, \answerNo{}, or \answerNA{}.
    \item[] Justification: We describe the model architecture in Section 4, the training objective in Section 4.2 and Appendix A.2.3, full hyperparameter settings in Appendix C.2.1 (Table 12), the noise schedule and time-dependent weighting in Appendix C.2.2, and the evaluation protocol in Section 5.1 and Appendix C.3. Code is publicly available at the project page linked in the abstract.
    \item[] Guidelines:
    \begin{itemize}
        \item The answer \answerNA{} means that the paper does not include experiments.
        \item If the paper includes experiments, a \answerNo{} answer to this question will not be perceived well by the reviewers: Making the paper reproducible is important, regardless of whether the code and data are provided or not.
        \item If the contribution is a dataset and\slash or model, the authors should describe the steps taken to make their results reproducible or verifiable. 
        \item Depending on the contribution, reproducibility can be accomplished in various ways. For example, if the contribution is a novel architecture, describing the architecture fully might suffice, or if the contribution is a specific model and empirical evaluation, it may be necessary to either make it possible for others to replicate the model with the same dataset, or provide access to the model. In general. releasing code and data is often one good way to accomplish this, but reproducibility can also be provided via detailed instructions for how to replicate the results, access to a hosted model (e.g., in the case of a large language model), releasing of a model checkpoint, or other means that are appropriate to the research performed.
        \item While NeurIPS does not require releasing code, the conference does require all submissions to provide some reasonable avenue for reproducibility, which may depend on the nature of the contribution. For example
        \begin{enumerate}
            \item If the contribution is primarily a new algorithm, the paper should make it clear how to reproduce that algorithm.
            \item If the contribution is primarily a new model architecture, the paper should describe the architecture clearly and fully.
            \item If the contribution is a new model (e.g., a large language model), then there should either be a way to access this model for reproducing the results or a way to reproduce the model (e.g., with an open-source dataset or instructions for how to construct the dataset).
            \item We recognize that reproducibility may be tricky in some cases, in which case authors are welcome to describe the particular way they provide for reproducibility. In the case of closed-source models, it may be that access to the model is limited in some way (e.g., to registered users), but it should be possible for other researchers to have some path to reproducing or verifying the results.
        \end{enumerate}
    \end{itemize}

\item {\bf Open access to data and code}
    \item[] Question: Does the paper provide open access to the data and code, with sufficient instructions to faithfully reproduce the main experimental results, as described in supplemental material?
    \item[] Answer: \answerYes{} % Replace by \answerYes{}, \answerNo{}, or \answerNA{}.
    \item[] Justification: Code and instructions for reproducing the main results are publicly available at https://dongbeank.github.io/MDTIM/. All datasets used (Energy, ETTh, Weather, Sine, PhysioNet 2012) are publicly available; URLs are provided in Appendix C.1.
    \item[] Guidelines:
    \begin{itemize}
        \item The answer \answerNA{} means that paper does not include experiments requiring code.
        \item Please see the NeurIPS code and data submission guidelines (\url{https://neurips.cc/public/guides/CodeSubmissionPolicy}) for more details.
        \item While we encourage the release of code and data, we understand that this might not be possible, so \answerNo{} is an acceptable answer. Papers cannot be rejected simply for not including code, unless this is central to the contribution (e.g., for a new open-source benchmark).
        \item The instructions should contain the exact command and environment needed to run to reproduce the results. See the NeurIPS code and data submission guidelines (\url{https://neurips.cc/public/guides/CodeSubmissionPolicy}) for more details.
        \item The authors should provide instructions on data access and preparation, including how to access the raw data, preprocessed data, intermediate data, and generated data, etc.
        \item The authors should provide scripts to reproduce all experimental results for the new proposed method and baselines. If only a subset of experiments are reproducible, they should state which ones are omitted from the script and why.
        \item At submission time, to preserve anonymity, the authors should release anonymized versions (if applicable).
        \item Providing as much information as possible in supplemental material (appended to the paper) is recommended, but including URLs to data and code is permitted.
    \end{itemize}

\item {\bf Experimental setting/details}
    \item[] Question: Does the paper specify all the training and test details (e.g., data splits, hyperparameters, how they were chosen, type of optimizer) necessary to understand the results?
    \item[] Answer: \answerYes{} % Replace by \answerYes{}, \answerNo{}, or \answerNA{}.
    \item[] Justification: Training and evaluation details are provided in Section 5.1 and Appendix C.2. Appendix C.2.1 (Table 12) lists the full hyperparameter configuration, including hidden size, number of layers, dropout, vocabulary size, batch size, learning rate, scheduler, warmup steps, gradient clipping, EMA decay, and FFT loss weight.
    \item[] Guidelines:
    \begin{itemize}
        \item The answer \answerNA{} means that the paper does not include experiments.
        \item The experimental setting should be presented in the core of the paper to a level of detail that is necessary to appreciate the results and make sense of them.
        \item The full details can be provided either with the code, in appendix, or as supplemental material.
    \end{itemize}

\item {\bf Experiment statistical significance}
    \item[] Question: Does the paper report error bars suitably and correctly defined or other appropriate information about the statistical significance of the experiments?
    \item[] Answer: \answerYes{} % Replace by \answerYes{}, \answerNo{}, or \answerNA{}.
    \item[] Justification: All main results are averaged over three random mask seeds (Section~\ref{sec:experiments}). Appendix~\ref{app:seeds} additionally reports the mean and standard deviation over three independently trained models (training seeds 2024, 2025, 2026) under each mask seed, for every dataset, missing pattern, and missing rate. The standard deviation is at or below 0.001 in every setting, well below the margins to the baselines.
    \item[] Guidelines:
    \begin{itemize}
        \item The answer \answerNA{} means that the paper does not include experiments.
        \item The authors should answer \answerYes{} if the results are accompanied by error bars, confidence intervals, or statistical significance tests, at least for the experiments that support the main claims of the paper.
        \item The factors of variability that the error bars are capturing should be clearly stated (for example, train/test split, initialization, random drawing of some parameter, or overall run with given experimental conditions).
        \item The method for calculating the error bars should be explained (closed form formula, call to a library function, bootstrap, etc.)
        \item The assumptions made should be given (e.g., Normally distributed errors).
        \item It should be clear whether the error bar is the standard deviation or the standard error of the mean.
        \item It is OK to report 1-sigma error bars, but one should state it. The authors should preferably report a 2-sigma error bar than state that they have a 96\% CI, if the hypothesis of Normality of errors is not verified.
        \item For asymmetric distributions, the authors should be careful not to show in tables or figures symmetric error bars that would yield results that are out of range (e.g., negative error rates).
        \item If error bars are reported in tables or plots, the authors should explain in the text how they were calculated and reference the corresponding figures or tables in the text.
    \end{itemize}

\item {\bf Experiments compute resources}
    \item[] Question: For each experiment, does the paper provide sufficient information on the computer resources (type of compute workers, memory, time of execution) needed to reproduce the experiments?
    \item[] Answer: \answerYes{} % Replace by \answerYes{}, \answerNo{}, or \answerNA{}.
    \item[] Justification: Compute resources are described in Appendix C.2: all experiments are conducted with dual Intel Xeon Gold 6444Y CPUs and a single NVIDIA H100 PCIe GPU (80GB). Inference time is reported in Table 5.
    \item[] Guidelines:
    \begin{itemize}
        \item The answer \answerNA{} means that the paper does not include experiments.
        \item The paper should indicate the type of compute workers CPU or GPU, internal cluster, or cloud provider, including relevant memory and storage.
        \item The paper should provide the amount of compute required for each of the individual experimental runs as well as estimate the total compute. 
        \item The paper should disclose whether the full research project required more compute than the experiments reported in the paper (e.g., preliminary or failed experiments that didn't make it into the paper). 
    \end{itemize}
    
\item {\bf Code of ethics}
    \item[] Question: Does the research conducted in the paper conform, in every respect, with the NeurIPS Code of Ethics \url{https://neurips.cc/public/EthicsGuidelines}?
    \item[] Answer: \answerYes{} % Replace by \answerYes{}, \answerNo{}, or \answerNA{}.
    \item[] Justification: Our research conforms to the NeurIPS Code of Ethics. The work uses publicly available benchmark datasets and does not involve human subjects, sensitive data, or applications with foreseeable harm.
    \item[] Guidelines:
    \begin{itemize}
        \item The answer \answerNA{} means that the authors have not reviewed the NeurIPS Code of Ethics.
        \item If the authors answer \answerNo, they should explain the special circumstances that require a deviation from the Code of Ethics.
        \item The authors should make sure to preserve anonymity (e.g., if there is a special consideration due to laws or regulations in their jurisdiction).
    \end{itemize}

\item {\bf Broader impacts}
    \item[] Question: Does the paper discuss both potential positive societal impacts and negative societal impacts of the work performed?
    \item[] Answer: \answerNA{} % Replace by \answerYes{}, \answerNo{}, or \answerNA{}.
    \item[] Justification: This work is foundational research on time series imputation methodology. While accurate imputation can benefit downstream applications such as healthcare and energy monitoring, the contribution is general-purpose and does not have a direct path to specific negative societal applications beyond those of generic time series modeling.
    \item[] Guidelines:
    \begin{itemize}
        \item The answer \answerNA{} means that there is no societal impact of the work performed.
        \item If the authors answer \answerNA{} or \answerNo, they should explain why their work has no societal impact or why the paper does not address societal impact.
        \item Examples of negative societal impacts include potential malicious or unintended uses (e.g., disinformation, generating fake profiles, surveillance), fairness considerations (e.g., deployment of technologies that could make decisions that unfairly impact specific groups), privacy considerations, and security considerations.
        \item The conference expects that many papers will be foundational research and not tied to particular applications, let alone deployments. However, if there is a direct path to any negative applications, the authors should point it out. For example, it is legitimate to point out that an improvement in the quality of generative models could be used to generate Deepfakes for disinformation. On the other hand, it is not needed to point out that a generic algorithm for optimizing neural networks could enable people to train models that generate Deepfakes faster.
        \item The authors should consider possible harms that could arise when the technology is being used as intended and functioning correctly, harms that could arise when the technology is being used as intended but gives incorrect results, and harms following from (intentional or unintentional) misuse of the technology.
        \item If there are negative societal impacts, the authors could also discuss possible mitigation strategies (e.g., gated release of models, providing defenses in addition to attacks, mechanisms for monitoring misuse, mechanisms to monitor how a system learns from feedback over time, improving the efficiency and accessibility of ML).
    \end{itemize}
    
\item {\bf Safeguards}
    \item[] Question: Does the paper describe safeguards that have been put in place for responsible release of data or models that have a high risk for misuse (e.g., pre-trained language models, image generators, or scraped datasets)?
    \item[] Answer: \answerNA{} % Replace by \answerYes{}, \answerNo{}, or \answerNA{}.
    \item[] Justification: The proposed model is a time series imputation method trained on public benchmark datasets and does not pose risks of misuse such as those associated with generative language or image models.
    \item[] Guidelines:
    \begin{itemize}
        \item The answer \answerNA{} means that the paper poses no such risks.
        \item Released models that have a high risk for misuse or dual-use should be released with necessary safeguards to allow for controlled use of the model, for example by requiring that users adhere to usage guidelines or restrictions to access the model or implementing safety filters. 
        \item Datasets that have been scraped from the Internet could pose safety risks. The authors should describe how they avoided releasing unsafe images.
        \item We recognize that providing effective safeguards is challenging, and many papers do not require this, but we encourage authors to take this into account and make a best faith effort.
    \end{itemize}

\item {\bf Licenses for existing assets}
    \item[] Question: Are the creators or original owners of assets (e.g., code, data, models), used in the paper, properly credited and are the license and terms of use explicitly mentioned and properly respected?
    \item[] Answer: \answerYes{} % Replace by \answerYes{}, \answerNo{}, or \answerNA{}.
    \item[] Justification: All datasets and baseline models are cited with their original publications. Dataset URLs are provided in Appendix C.1 footnotes. Baseline implementations follow the codebase of Du et al.~\cite{du2024tsi}, also referenced in Appendix C.1.
    \item[] Guidelines:
    \begin{itemize}
        \item The answer \answerNA{} means that the paper does not use existing assets.
        \item The authors should cite the original paper that produced the code package or dataset.
        \item The authors should state which version of the asset is used and, if possible, include a URL.
        \item The name of the license (e.g., CC-BY 4.0) should be included for each asset.
        \item For scraped data from a particular source (e.g., website), the copyright and terms of service of that source should be provided.
        \item If assets are released, the license, copyright information, and terms of use in the package should be provided. For popular datasets, \url{paperswithcode.com/datasets} has curated licenses for some datasets. Their licensing guide can help determine the license of a dataset.
        \item For existing datasets that are re-packaged, both the original license and the license of the derived asset (if it has changed) should be provided.
        \item If this information is not available online, the authors are encouraged to reach out to the asset's creators.
    \end{itemize}

\item {\bf New assets}
    \item[] Question: Are new assets introduced in the paper well documented and is the documentation provided alongside the assets?
    \item[] Answer: \answerYes{} % Replace by \answerYes{}, \answerNo{}, or \answerNA{}.
    \item[] Justification: Code for MDTIM is publicly released at https://dongbeank.github.io/MDTIM/ with documentation describing how to reproduce the experiments.
    \item[] Guidelines:
    \begin{itemize}
        \item The answer \answerNA{} means that the paper does not release new assets.
        \item Researchers should communicate the details of the dataset\slash code\slash model as part of their submissions via structured templates. This includes details about training, license, limitations, etc. 
        \item The paper should discuss whether and how consent was obtained from people whose asset is used.
        \item At submission time, remember to anonymize your assets (if applicable). You can either create an anonymized URL or include an anonymized zip file.
    \end{itemize}

\item {\bf Crowdsourcing and research with human subjects}
    \item[] Question: For crowdsourcing experiments and research with human subjects, does the paper include the full text of instructions given to participants and screenshots, if applicable, as well as details about compensation (if any)? 
    \item[] Answer: \answerNA{} % Replace by \answerYes{}, \answerNo{}, or \answerNA{}.
    \item[] Justification: This work does not involve crowdsourcing or research with human subjects.
    \item[] Guidelines:
    \begin{itemize}
        \item The answer \answerNA{} means that the paper does not involve crowdsourcing nor research with human subjects.
        \item Including this information in the supplemental material is fine, but if the main contribution of the paper involves human subjects, then as much detail as possible should be included in the main paper. 
        \item According to the NeurIPS Code of Ethics, workers involved in data collection, curation, or other labor should be paid at least the minimum wage in the country of the data collector. 
    \end{itemize}

\item {\bf Institutional review board (IRB) approvals or equivalent for research with human subjects}
    \item[] Question: Does the paper describe potential risks incurred by study participants, whether such risks were disclosed to the subjects, and whether Institutional Review Board (IRB) approvals (or an equivalent approval/review based on the requirements of your country or institution) were obtained?
    \item[] Answer: \answerNA{} % Replace by \answerYes{}, \answerNo{}, or \answerNA{}.
    \item[] Justification: This work does not involve research with human subjects. The PhysioNet 2012 dataset is a publicly released benchmark with the original IRB approval handled by its providers.
    \item[] Guidelines:
    \begin{itemize}
        \item The answer \answerNA{} means that the paper does not involve crowdsourcing nor research with human subjects.
        \item Depending on the country in which research is conducted, IRB approval (or equivalent) may be required for any human subjects research. If you obtained IRB approval, you should clearly state this in the paper. 
        \item We recognize that the procedures for this may vary significantly between institutions and locations, and we expect authors to adhere to the NeurIPS Code of Ethics and the guidelines for their institution. 
        \item For initial submissions, do not include any information that would break anonymity (if applicable), such as the institution conducting the review.
    \end{itemize}

\item {\bf Declaration of LLM usage}
    \item[] Question: Does the paper describe the usage of LLMs if it is an important, original, or non-standard component of the core methods in this research? Note that if the LLM is used only for writing, editing, or formatting purposes and does \emph{not} impact the core methodology, scientific rigor, or originality of the research, declaration is not required.
    %this research? 
    \item[] Answer: \answerNA{} % Replace by \answerYes{}, \answerNo{}, or \answerNA{}.
    \item[] Justification: LLMs are not used as a component of the core methodology in this research.
    \item[] Guidelines:
    \begin{itemize}
        \item The answer \answerNA{} means that the core method development in this research does not involve LLMs as any important, original, or non-standard components.
        \item Please refer to our LLM policy in the NeurIPS handbook for what should or should not be described.
    \end{itemize}

\end{enumerate}

\end{document}